\documentclass{ieeeaccess}

\usepackage{bm}
\usepackage{url}

\usepackage{cite}
\usepackage{amsmath,amssymb,amsfonts}
\usepackage{algorithmic}
\usepackage{graphicx}
\usepackage{textcomp}

\usepackage{comment}

\usepackage{booktabs}
\usepackage{multirow}

\usepackage[colorlinks]{hyperref}

\usepackage{caption, subcaption}
\def\BibTeX{{\rm B\kern-.05em{\sc i\kern-.025em b}\kern-.08em
    T\kern-.1667em\lower.7ex\hbox{E}\kern-.125emX}}

\newcommand{\argmax}{\mathop{\mathrm{argmax}}\limits}

\begin{document}
\history{Received May 18, 2021, accepted Jun 21, 2021, date of publication xxxx 00, 0000, date of current version xxxx 00, 0000.}
\doi{10.1109/ACCESS.2021.3093456}

\title{
    Attention Meets Perturbations: \\ 
    Robust and Interpretable Attention \\ 
    with Adversarial Training
}
\author{
    \uppercase{Shunsuke Kitada\authorrefmark{1}, and Hitoshi Iyatomi}\authorrefmark{1}\IEEEmembership{Member, IEEE}
}
\address[1]{
    Hosei University, Graduate School of Science and Engineering, Department of Applied Informatics, Tokyo 1883-8584, Japan
}
\tfootnote{
This work was supported by JSPS KAKENHI Grant Number 21J14143.
}

\markboth
{S. Kitada and H. Iyatomi: Attention Meets Perturbations: Robust and Interpretable Attention with Adversarial Training}
{S. Kitada and H. Iyatomi: Attention Meets Perturbations: Robust and Interpretable Attention with Adversarial Training}

\corresp{Corresponding author: Shunsuke Kitada (e-mail: shunsuke.kitada.8y@stu.hosei.ac.jp).}

\begin{abstract}
    Although attention mechanisms have been applied to a variety of deep learning models and have been shown to improve the prediction performance, it has been reported to be vulnerable to perturbations to the mechanism. 
To overcome the vulnerability to perturbations in the mechanism, we are inspired by adversarial training (AT), which is a powerful regularization technique for enhancing the robustness of the models.
In this paper, we propose a general training technique for natural language processing tasks, including AT for attention (Attention AT) and more interpretable AT for attention (Attention iAT). 
The proposed techniques improved the prediction performance and the model interpretability by exploiting the mechanisms with AT. 
In particular, Attention iAT boosts those advantages by introducing adversarial perturbation, which enhances the difference in the attention of the sentences. 
Evaluation experiments with ten open datasets revealed that AT for attention mechanisms, especially Attention iAT, demonstrated (1) the best performance in nine out of ten tasks and (2) more interpretable attention (i.e., the resulting attention correlated more strongly with gradient-based word importance) for all tasks. 
Additionally, the proposed techniques are (3) much less dependent on perturbation size in AT. 

\end{abstract}


\titlepgskip=-15pt

\maketitle

\section{Introduction}\label{sec:introduction}
\IEEEPARstart{A}{ttention} mechanisms~\cite{bahdanau2014neural} are widely applied in natural language processing (NLP) field through deep neural networks (DNNs).
As the effectiveness of attention mechanisms became apparent in various tasks~\cite{lin2017structured, wang2016attention, he2016character, parikh2016decomposable, luong2015effective, rush2015neural}, they were applied not only to recurrent neural networks (RNNs) but also to convolutional neural networks (CNNs). 
Moreover, Transformers~\cite{vaswani2017attention} which make proactive use of attention mechanisms have also achieved excellent results.
However, it has been pointed out that DNN models tend to be locally unstable, and even tiny perturbations to the original inputs~\cite{szegedy2013intriguing} or attention mechanisms can mislead the models~\cite{jain2019attention}.
Specifically, Jain and Wallace~\cite{jain2019attention} used a practical bi-directional RNN (BiRNN) model to investigate the effect of attention mechanisms and reported that learned attention weights based on the model are vulnerable to perturbations.\footnote{In Jain and Wallace~\cite{jain2019attention}, the vulnerability of attention mechanisms to perturbations is confirmed with an RNN-based model~\cite{jain2019attention}. In this paper, we focus on the model, and Transformer~\cite{vaswani2017attention}-based model such as BERT~\cite{liu2016delving} and their successor models~\cite{liu2019roberta, lan2020albert} will be future work.}

The Transformer~\cite{vaswani2017attention} and its follow-up models~\cite{liu2019roberta, lan2020albert} have self-attention mechanisms that estimate the relationship of each word in the sentence.
These models take advantage of the effect of the mechanisms and have shown promising performances.
Thus, there is no doubt that the effect of the mechanisms is extremely large.
However, they are not easy to train, as they require huge amounts of GPU memory to maintain the weights of the model.
Recently, there have been proposals to reduce memory consumption~\cite{tay2020efficient}, and we acknowledge the advantages of the models.
On the other hand, the application of attention mechanisms to DNN models, such as RNN and CNN models, which have been widely used and do not require relatively high training requirements, has not been sufficiently studied.

In this paper, we focus on improving the robustness of commonly used BiRNN models (as described detail in Section~\ref{sec:common_model_architecture}) to perturbations in the attention mechanisms.
Furthermore, we demonstrate that the result of overcoming the vulnerability of the attention mechanisms is an improvement in the prediction performance and model interpretability.

To tackle the models' vulnerability to perturbation, Goodfellow et al.~\cite{goodfellow2014explaining} proposed adversarial training (AT) that increases robustness by adding adversarial perturbations to the input and the training technique forcing the model to address its difficulties. 
Previous studies~\cite{goodfellow2014explaining,shaham2018understanding} in the image recognition field have theoretically explained the regularization effect of AT and shown that it improves the robustness of the model for unseen images.

\begin{figure}
    \centering
    \includegraphics[width=\linewidth]{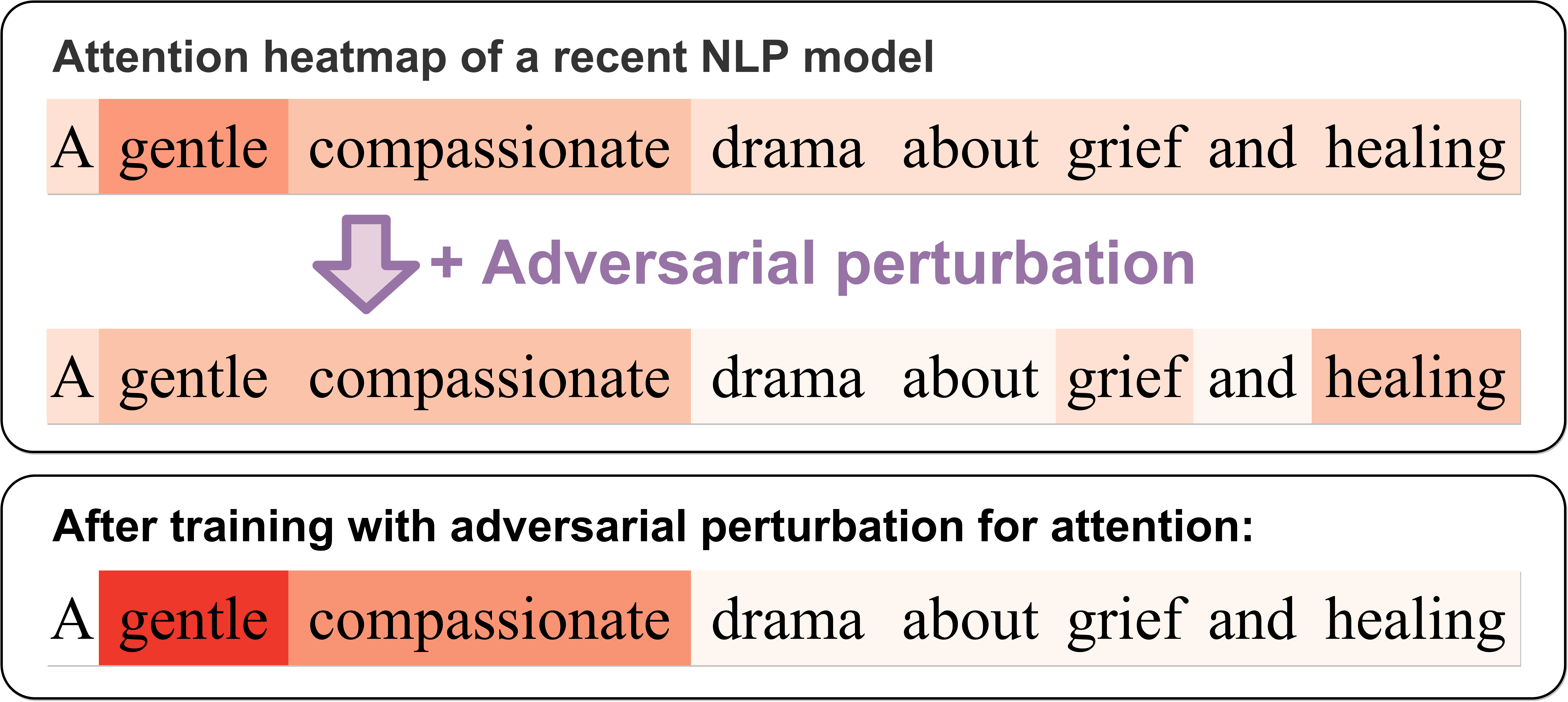}
    \caption{An example of an attention heatmap for a BiRNN model with attention mechanisms and the model with attention mechanisms trained with adversarial training from the Stanford Sentiment Treebank (SST)~\cite{socher2013recursive}. The proposed adversarial training for attention mechanisms helps the model learn cleaner attention.}
    \label{fig:figure1}
\end{figure}

AT is also widely used in the NLP field as a powerful regularization technique~\cite{miyato2016adversarial, sato2018interpretable,yasunaga2018robust,wang2018robust}.
In pioneering work, Miyato et al.~\cite{miyato2016adversarial} proposed a simple yet effective technique to improving the text classification performance by applying AT to a word embedding space.
Later, interpretable AT (iAT) was proposed to increase the interpretability of the model by restricting the direction of the perturbations to existing words in the word embedding space~\cite{sato2018interpretable}.
The attention weight of each word is considered an indicator of the importance of each word~\cite{li2016understanding}, and thus, in terms of interpretability, we assume that the weight is considered a higher-order feature than the word embedding.
Therefore, AT for attention mechanisms that adds an adversarial perturbation to deceive the attention mechanisms is expected to be more effective than AT for word embedding.

From motivations above, we propose a new general training technique for attention mechanisms based on AT, called \textit{adversarial training for attention} (Attention AT) and \textit{more interpretable adversarial training for attention} (Attention iAT). 
The proposed techniques are the first attempt to employ AT for attention mechanisms.
The proposed Attention AT/iAT is expected to improve the robustness and the interpretability of the model by appropriately overcoming the adversarial perturbations to attention mechanisms~\cite{tsipras2019robustness, itazuri2019adversarially, zhang2019interpreting}.
Because our proposed AT techniques for attention mechanisms is model-independent and a general technique, it can be applied to various DNN models (e.g., RNN and CNN) with attention mechanisms.
Our technique can also be applied to any similarity functions for attention mechanisms, e.g, additive function~\cite{bahdanau2014neural} and scaled dot-product function~\cite{vaswani2017attention}, which is famous for calculating the similarity in attention mechanisms.

To demonstrate the effects of these techniques, we evaluated them compared to several other state-of-the-art AT-based techniques~\cite{miyato2016adversarial, sato2018interpretable} with ten common datasets for different NLP tasks. 
These datasets included binary classification (BC), question answering (QA), and natural language inference (NLI).
We also evaluated how the attention weights obtained through the proposed AT technique agreed with the word importance calculated by the gradients~\cite{simonyan2013deep}. 
Evaluating the proposed techniques, we obtained the following findings concerning AT for attention mechanisms in NLP:
\begin{itemize}
\item AT for attention mechanisms improves the prediction performance of various NLP tasks.
\item AT for attention mechanisms helps the model learn cleaner attention (as shown in Figure~\ref{fig:figure1}) and demonstrates a stronger correlation with the word importance calculated from the model gradients.
\item The proposed training techniques are much less independent concerning perturbation size in AT.
\end{itemize}
Especially, our Attention iAT demonstrated the best performance in nine out of ten tasks and more interpretable attention, i.e., resulting attention weight correlated more strongly with the gradient-based word importance~\cite{simonyan2013deep}.
The implementation required to reproduce these techniques and the evaluation experiments are available on GitHub.\footnote{\url{https://github.com/shunk031/attention-meets-perturbation}}

\section{Related Work}
\subsection{Attention Mechanisms}
Attention mechanisms were introduced by Bahdanau et al.~\cite{bahdanau2014neural} for the task of machine translation.
Today, these mechanisms contribute to improving the prediction performance of various tasks in the NLP field, such as sentence-level classification~\cite{lin2017structured}, sentiment analysis~\cite{wang2016attention}, question answering~\cite{he2016character}, and natural language inference~\cite{parikh2016decomposable}.
There are a wide variety of attention mechanisms; for instance, additive~\cite{bahdanau2014neural} and scaled dot-product~\cite{vaswani2017attention} functions are used as similarity functions.

Attention weights are often claimed to offer insights into the inner workings of DNNs~\cite{li2016understanding}.
However, Jain and Wallace~\cite{jain2019attention} reported that learned attention weights are often uncorrelated with the word importance calculated through the gradient-based method~\cite{simonyan2013deep}, and perturbations interfere with interpretation.
In this paper, we demonstrate that AT for attention mechanisms can mitigate these issues.

\begin{figure*}
    \begin{minipage}{0.37\textwidth}
        \centering
        \includegraphics[width=\linewidth]{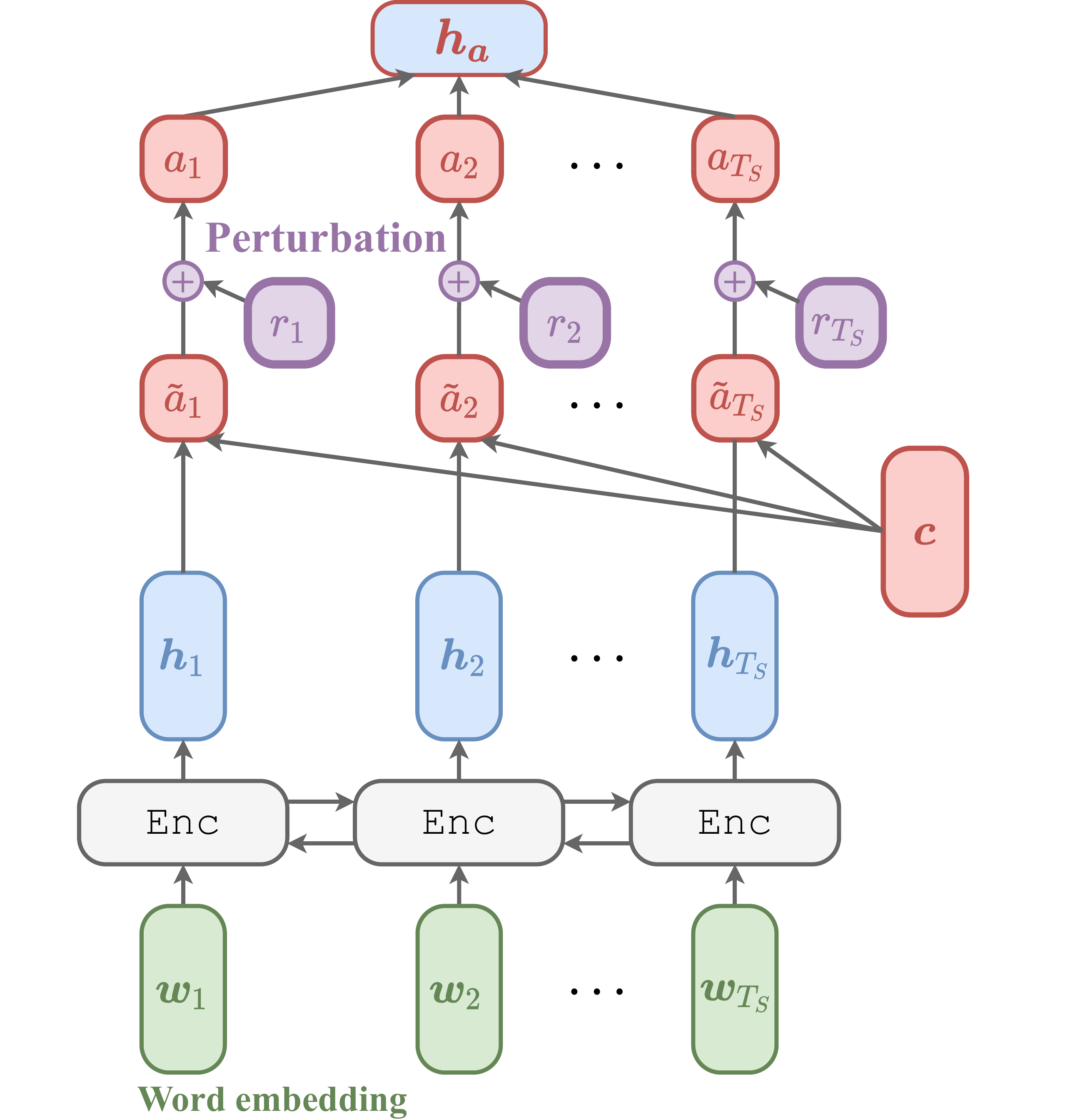}
        \subcaption{Single sequence model}
        \label{fig:model_for_bc}
    \end{minipage}
    \begin{minipage}{0.63\textwidth}
        \centering
        \includegraphics[width=\linewidth]{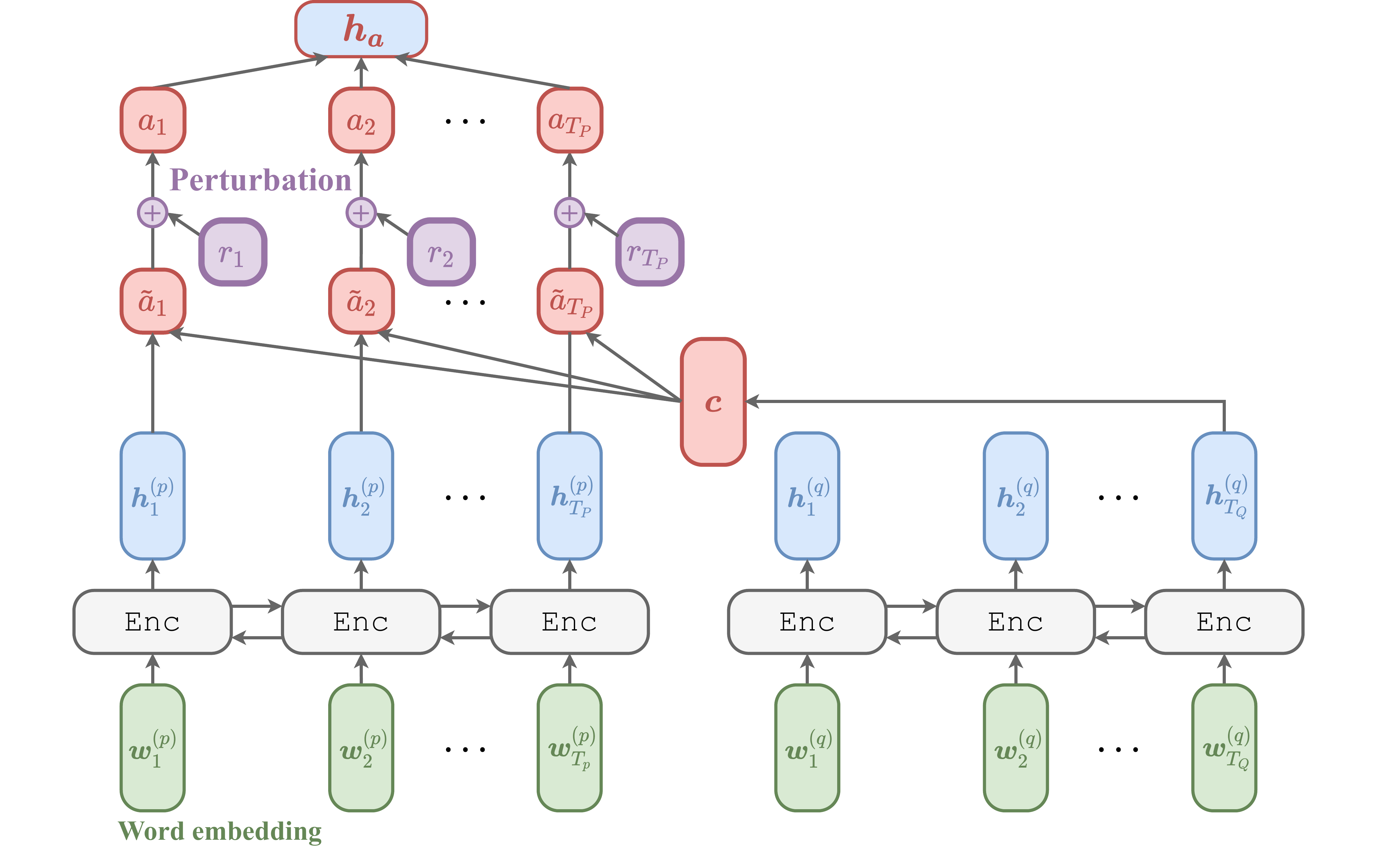}
        \subcaption{Pair sequence model}
        \label{fig:model_for_qa_nli}
    \end{minipage}
    \caption{Illustration of the common model architectures to apply the proposed training technique: (a) a single sequence model for the binary classification task and (b) a pair sequence model for question answering and natural language inference tasks. In (a), the input of the model is word embeddings, $\{\bm{w}_1, \cdots, \bm{w}_{T_S}\}$ associated with the input sentence $X_S$. In (b), the inputs are word embeddings $\{\bm{w}_{1}^{(p)}, \cdots, \bm{w}_{T_P}^{(p)}\}$ and $\{\bm{w}_{1}^{(q)}, \cdots, \bm{w}_{T_Q}^{(q)}\}$ from two input sequences, $X_P$ and $X_Q$, respectively. These inputs are encoded into hidden states through a bi-directional encoder (Enc). In conventional models, the worst-case perturbation $\bm{r}$ is added to the word embeddings. In our adversarial training for attention mechanisms, we compute and add the $\bm{r}$ to attention $\bm{a}$ to improve the prediction performance and the interpretability of the model.}
    \label{fig:model_overview}
\end{figure*}

\subsection{Adversarial Training}
AT~\cite{szegedy2013intriguing, goodfellow2014explaining, wang2016theoretical} is a powerful regularization technique that has been primarily explored in the field of image recognition to improve the robustness of models for input perturbations.
In the NLP field, AT has been applied to various tasks by extending the concept of adversarial perturbations, e.g., text classification~\cite{miyato2016adversarial, sato2018interpretable}, part-of-speech tagging~\cite{yasunaga2018robust}, and machine reading comprehension~\cite{wang2018robust, liu2020robust}.
As mentioned earlier, these techniques apply AT for word embedding.
Other AT-based techniques for the NLP tasks include those related to parameter updating~\cite{barham2019interpretable} and generative adversarial network (GAN)-based retrieval-enhancement method~\cite{zhu2019retrieval}.
Our proposal is an adversarial training technique for attention mechanisms and is different from these methods.

Miyato et al.~\cite{miyato2016adversarial} proposed Word AT, a technique that applied AT to the word embeddings.
The adversarial perturbations are generated according to the back-propagation gradients.
These perturbations are expected to regularize the model.
Since then, Sato et al.~\cite{sato2018interpretable} proposed Word iAT, and it has been known to achieve almost the same performance as Word AT that does not expect interpretability~\cite{sato2018interpretable}.
The Word iAT technique aims to increase the model's interpretability by determining the perturbation's direction so that it is closer to other word embeddings in the vocabulary.
Both reports demonstrated improved task performance via AT. 
However, the specific effect of AT on attention mechanisms has yet to be investigated.
In this paper, we aim to address this issue by providing analyses of the effects of AT for attention mechanisms using various NLP tasks.

AT is considered to be related to other regularization techniques (e.g., dropout~\cite{srivastava2014dropout}, batch normalization~\cite{ioffe2015batch}).
Specifically, dropout can be considered a kind of noise addition.
Word dropout~\cite{iyyer2015deep} and character dropout~\cite{shimada2016document}, known as wildcard training, are variants for NLP tasks.
These techniques can be considered random noise for the target task.
In contrast, AT has been demonstrated to be effective because it creates particularly vulnerable perturbations that the model is trained to overcome~\cite{goodfellow2014explaining}.

It has been reported that DNN models that introduce adversarial training to overcome adversarial perturbations capture human-like features~\cite{tsipras2019robustness, itazuri2019adversarially, zhang2019interpreting}.
These features help to make the prediction of DNN models easier to interpret for humans.
In this paper, we demonstrate that the proposed AT to attention mechanisms provides cleaner attention that is more easily interpreted by humans.

\section{Common Model Architecture}\label{sec:common_model_architecture}
Our goal is to improve the performance of NLP models (i.e., predictability and interpretability) by aiming at the robustness of the attention mechanisms.
To demonstrate the effectiveness of exploiting AT to attention for words, we adopted the BiRNN-based model used by Jain and Wallace~\cite{jain2019attention} as our common model architecture and set their performance as our performance baseline, as described in Section~\ref{sec:introduction}.
This is because they performed extensive experiments across a variety of public NLP tasks to investigate the effect of attention mechanisms, and their model has demonstrated desirable prediction performance.
However, the attention mechanism in the model has been reported to be vulnerable to perturbations.

Based on the model of Jain and Wallace~\cite{jain2019attention}, we investigated three common NLP tasks, BC, QA, and NLI.
Because Jain and Wallace~\cite{jain2019attention} considered the same tasks.
A BC task is a single sequence task that takes one input text, while QA and NLI tasks are pair sequence tasks that take two input sequences.
Then, we defined two base models, a single sequence model and pair sequence model, for those tasks, as shown in Figure~\ref{fig:model_overview}.

\subsection{Model with Attention Mechanisms for Single Sequence Task}
For a single sequence task, such as the BC task, the input is a word sequence of one-hot encoding $X_S = (\bm{x}_1, \bm{x}_2, \cdots, \bm{x}_{T_S}) \in \mathbb{R}^{T_S \times |V|}$, where $T_S$ and $|V|$ are the number of words in the sentence and vocabulary size.
We introduce the following short notation for the sequence $(\bm{x}_1, \bm{x}_2, \cdots, \bm{x}_{T_S})$ as $(\bm{x}_t)_{t=1}^{T_S}$.
Let $\bm{w}_t \in \mathbb{R}^{d}$ be a $d$-dimensional word embedding that corresponds to $\bm{x}_t$.
We represent each word with the word embeddings to obtain $(\bm{w}_t)_{t=1}^{T_S} \in \mathbb{R}^{T_S \times d}$.
Next, we use the BiRNN encoder $\mathbf{Enc}$ to obtain $m$-dimensional hidden states $\bm{h}_t$:
\begin{equation}
    \bm{h}_t = \mathbf{Enc}(\bm{w}_t, \bm{h}_{t-1}),
\end{equation}
where $\bm{h}_0$ is the initial hidden state and is regarded as a zero vector.
Next, we use the additive formulation of attention mechanisms proposed by Bahdanau et al.~\cite{bahdanau2014neural} to compute the attention score for the $t$-th word $\tilde{a}_t$, defined as:
\begin{equation}
    \tilde{a}_t = \bm{c}^{\top} \mathrm{tanh}(\bm{W}\bm{h}_t + \bm{b}),\label{eq:attention_scalar_bc}
\end{equation}
where $\bm{W} \in \mathbb{R}^{d' \times m}$ and $\bm{b}, \bm{c} \in \mathbb{R}^{d'}$ are the parameters of the model.
Then, from the attention scores $\tilde{\bm{a}} = (\tilde{a}_t)_{t=1}^{T_S}$, the attention weights $\bm{a} = (a_t)_{t=1}^{T_S}$ for all words are computed as
\begin{equation}
    \bm{a} = \mathrm{softmax}(\tilde{\bm{a}}).\label{eq:attention_weight}
\end{equation}
The weighted instance representation $\bm{h}_{\bm{a}}$ is calculated as
\begin{equation}
    \bm{h}_{\bm{a}} = \sum_{t=1}^{T_S} a_t \bm{h}_t.
\end{equation}
Finally, $\bm{h}_{\bm{a}}$ is fed to a dense layer \textbf{Dec}, and the output activation function is then used to obtain the predictions:
\begin{equation}
    \hat{\bm{y}} = \sigma(\mathbf{Dec}(\bm{h}_{\bm{\bm{a}}})) \in \mathbb{R}^{|\bm{y}|},\label{eq:prediction}
\end{equation}
where $\sigma$ is a sigmoid function, and $|\bm{y}|$ is the label set size.

\subsection{Model with Attention Mechanisms for Pair Sequence Task}
For a pair sequence task, such as the QA and NLI tasks, the input is $X_P = (\bm{x}_t^{(p)})_{t=1}^{T_P} \in \mathbb{R}^{T_P \times |V|}$ and $X_Q = (\bm{x}_t^{(q)})_{t=1}^{T_Q} \in \mathbb{R}^{T_Q \times |V|}$.
$T_P$ and $T_Q$ are the number of words in each sentence.
$X_P$ and $X_Q$ represent the paragraph and question in the QA and the hypothesis and premise in the NLI.
We used two separate BiRNN encoders to obtain the hidden states $\bm{h}_{t}^{(p)} \in \mathbb{R}^m$ and $\bm{h}_{t}^{(q)} \in \mathbb{R}^m$:
\begin{equation}
    \bm{h}_{t}^{(p)} = \mathbf{Enc}_P(\bm{w}_{t}^{(p)}, \bm{h}_{t-1}^{(p)}); ~~ \bm{h}_{t}^{(q)} = \mathbf{Enc}_Q(\bm{w}_{t}^{(q)}, \bm{h}_{t-1}^{(q)}),
\end{equation}
where $\bm{h}_{0}^{(p)}$ and $\bm{h}_{0}^{(q)}$ are the initial hidden states and are regarded as zero vectors.
Next, we computed the attention weight $\tilde{a}_t$ of each word of $X_P$ as:
\begin{equation}
    \tilde{a}_t = \bm{c}^{\top} \mathrm{tanh}(\bm{W}_1\bm{h}_t^{(p)} + \bm{W}_2\bm{h}_{T_Q}^{(q)} + \bm{b}),
\end{equation}
where $\bm{W}_1 \in \mathbb{R}^{'d \times m}$ and $\bm{W}_2 \in \mathbb{R}^{d' \times m}$ denote the projection matrices, and $\bm{b}, \bm{c} \in \mathbb{R}^{d'}$ are the parameter vectors.
Similar to Eq.~\ref{eq:attention_weight}, the attention weight $a_t$ can be calculated from $\tilde{a}_t$.
The presentation is obtained from a sum of words in $X_P$.
\begin{equation}
    \bm{h}_{\bm{a}} = \sum_{t=1}^{T_P} a_t \bm{h}_t^{(p)}
\end{equation}
is fed to a \textbf{Dec}, and then a softmax function is used as $\sigma$ to obtain the prediction (in the same manner as in Eq.~\ref{eq:prediction}).

\subsection{Training Model with Attention Mechanisms}

Let $X_{\tilde{\bm{a}}}$ be an input sequence with attention score $\tilde{\bm{a}}$, where $\tilde{\bm{a}}$ is a concatenated attention score for all $t$. 
We model the conditional probability of the class $\bm{y}$ as $p(\bm{y}|X_{\tilde{\bm{a}}}; \bm{\theta})$, where $\bm{\theta}$ represents all model parameters.
For training the model, we minimize the following negative log likelihood as a loss function with respect to the model parameters:
\begin{equation}
    \mathcal{L}(X_{\tilde{\bm{a}}}, \bm{y}; \bm{\theta}) = - \log{p(\bm{y}|X_{\tilde{\bm{a}}}; \bm{\theta})}.
\end{equation}

\section{Adversarial Training for Attention Mechanisms}
The main contribution of this paper is to explore the idea of employing AT for attention mechanisms.
In this paper, we propose a new training technique for attention mechanisms based on AT, called Attention AT and Attention iAT.
The proposed techniques aim to achieve better regularization effects and to provide better interpretation of attention in the sentence.
These techniques are the first application of AT to the attention in each word, which is expected to be more interpretable, with reference to AT for word embeddings~\cite{miyato2016adversarial} and a technique more focused on interpretability~\cite{sato2018interpretable}.
In this paper, we generate adversarial perturbations based on the model described in Section~\ref{sec:common_model_architecture}.

\subsection{Attention AT: Adversarial Training for Attention}\label{sec:method_attention_at}
We describe the proposed Attention AT, which features adversarial perturbations in the attention mechanisms rather than in the word embeddings~\cite{miyato2016adversarial, sato2018interpretable}.
The adversarial perturbation on the mechanisms is defined as the worst-case perturbation on attention mechanisms of a small bounded norm $\epsilon$ that maximizes loss function $\mathcal{L}$ of the current model:
\begin{equation}
    \bm{r}_{\tt AT} = \argmax_{\bm{r}: ||\bm{r}||_2 \le \epsilon} \mathcal{L}(X_{\tilde{\bm{a}} + \bm{r}}, \bm{y}; \hat{\bm{\theta}})\label{eq:argmax},
\end{equation}
where $X_{\tilde{\bm{a}} + \bm{r}}$ is the input sequence with attention score $\tilde{\bm{a}}$, its perturbation $\bm{r}$, $\bm{y}$ is the target output, and $\hat{\bm{\theta}}$ represents the current model parameters.
We apply the fast gradient method~\cite{liu2016delving, miyato2016adversarial}, i.e., first-order approximation to obtain an approximate worst-case perturbation of norm $\epsilon$, through a single gradient computation as follows:
\begin{equation}
    \bm{r}_{\tt AT} = \epsilon \frac{\bm{g}}{||\bm{g}||_2}, \mathrm{where}~ \bm{g} = \nabla_{\tilde{\bm{a}}} \mathcal{L}(X_{\tilde{\bm{a}}}, \bm{y}; \hat{\bm{\theta}}).\label{eq:fast_gradient_method}
\end{equation}
$\epsilon$ is a hyper-parameter to be determined using the validation dataset.
We find this $\bm{r}_{\tt AT}$ against the current model parameterized by $\hat{\bm{\theta}}$ at each training step and construct an adversarial perturbation for attention score $\tilde{\bm{a}}$:
\begin{equation}
    \tilde{\bm{a}}_{\tt adv} = \tilde{\bm{a}} + \bm{r}_{\tt AT}.\label{eq:alpha_plus_r_at}
\end{equation}

\subsection{Attention iAT: Interpretable Adversarial Training for Attention}
We describe the proposed Attention iAT for further boosting the prediction performance and the interpretability of NLP tasks.
Rather than utilizing AT to attention mechanisms (as described in Section~\ref{sec:method_attention_at}), Attention iAT effectively exploits differences in the attention to each word in a sentence for the training.
As a result, this technique provides cleaner attention in the sentence and improves the interpretability of the attention.
These effects contribute to improving the performance of various NLP tasks.

\begin{table*}[t]
\centering
\caption{Dataset statistics. We used the three most well-known NLP tasks for evaluation: binary classification (BC), question answering (QA), and natural language inference (NLI). We split the dataset into training, validation, and test sets. We performed preprocessing as shown at \texttt{\url{https://github.com/successar/AttentionExplanation}} in the same manner as Jain and Wallace~\cite{jain2019attention}. See the details in Appendix~\ref{sec:tasks_and_datasets}. Jain and Wallace~\cite{jain2019attention} split the dataset into only a training set and a test set, so we did not get the same result.}
%\resizebox{\textwidth}{!}{%
\begin{tabular}{@{}llllrrrrrr@{}}
\toprule
Task                                                                                   & \multicolumn{3}{l}{Dataset}                                                & \multicolumn{1}{l}{\# class} & \multicolumn{1}{l}{\# train} & \multicolumn{1}{l}{\# valid} & \multicolumn{1}{l}{\# test} & \multicolumn{1}{c}{\# vocab} & \multicolumn{1}{l}{Avg. \# words} \\ \cmidrule(r){1-1} \cmidrule(lr){2-4} \cmidrule(lr){5-5} \cmidrule(lr){6-8} \cmidrule(lr){9-9} \cmidrule(l){10-10}
\multirow{4}{*}{\begin{tabular}[c]{@{}l@{}}Binary \\ Classification (BC)\end{tabular}}      & \multicolumn{2}{l}{SST}        & \cite{socher2013recursive}                & 2                            & 6,920                        & 872                          & 1,821                       & 13,723                             & 18                               \\
                                                                                       & \multicolumn{2}{l}{IMDB}       & \cite{maas2011learning}                   & 2                            & 17,186                       & 4,294                        & 4,353                       & 12,485                             & 171                              \\
                                                                                       & \multicolumn{2}{l}{20News}     & \cite{lang1995newsweeder}                 & 2                            & 1,145                        & 278                          & 357                         & 5,986                              & 110                              \\
                                                                                       & \multicolumn{2}{l}{AGNews}     & \cite{zhang2015character}                 & 2                            & 51,000                       & 9,000                        & 3,800                       & 13,713                             & 35                               \\ \cmidrule(r){1-1} \cmidrule(lr){2-4} \cmidrule(lr){5-5} \cmidrule(lr){6-8} \cmidrule(lr){9-9} \cmidrule(l){10-10}
\multirow{4}{*}{\begin{tabular}[c]{@{}l@{}}Question \\ Answering (QA)\end{tabular}}         & \multicolumn{2}{l}{CNN news}        & \cite{hermann2015teaching}                & 584                          & 380,298                      & 3,924                        & 3,198                       & 70,192                             & 773                              \\
                                                                                       & \multirow{3}{*}{bAbI} & Task 1 &  & 6                            & 8,500                        & 1,500                        & 1,000                       & 22                                 & 39                               \\
                                                                                       &                       & Task 2 & \cite{weston2015towards}                                          & 6                            & 8,500                        & 1,500                        & 1,000                       & 36                                 & 98                               \\
                                                                                       &                       & Task 3 &                                           & 6                            & 8,500                        & 1,500                        & 1,000                       & 37                                 & 313                              \\ \cmidrule(r){1-1} \cmidrule(lr){2-4} \cmidrule(lr){5-5} \cmidrule(lr){6-8} \cmidrule(lr){9-9} \cmidrule(l){10-10}
\multirow{2}{*}{\begin{tabular}[c]{@{}l@{}}Natural Language \\ Inference (NLI)\end{tabular}} & \multicolumn{2}{l}{SNLI}       & \cite{bowman2015large}                    & 3                            & 549,367                      & 9,842                        & 9,824                       & 20,979                             & 22                               \\
                                                                                       & \multicolumn{2}{l}{Multi NLI}  & \cite{williams2017broad}                  & 3                            & 314,161                      & 78,541                       & 19,647                      & 53,112                             & 34                               \\ \bottomrule
\end{tabular}%
%}
\label{tab:dataset}
\end{table*}

In terms of formulation, the proposed Attention iAT is analogous to interpretable AT for word embeddings (Word iAT)~\cite{sato2018interpretable}, which increases the interpretability of AT for word embeddings in formulas.
However, the implications and effects for training the model are very different; in the proposed Attention iAT, the \textit{attention difference enhancement}, described later, enhances the difference in attention for each word.
The difference and its effect will be explained later in this section and discussed in Section~\ref{sec:discussion_attention_at_attention_iat_word_iat}.

Suppose $\tilde{a}_t$ denotes the attention score corresponding to the $t$-th position in the sentence.
We define the difference vector $\bm{d}_t$ as the difference between the attention to the $t$-th word $\tilde{a}_t$ in a sentence and the attention to any $k$-th word $\tilde{a}_k$:
\begin{equation}
    \bm{d}_t = (d_{t, k})_{k=1}^{T} = (\tilde{a}_t - \tilde{a}_k)_{k=1}^{T}.\label{eq:dist_attention_in_sentence}
\end{equation}
$T = T_S$ in single sequence task, and $T = T_P$ in a pair sequence task.
By normalizing the norm of the vector, we define a normalized difference vector of the attention for the $t$-th word:
\begin{equation}
    \tilde{\bm{d}}_t = \frac{\bm{d}_t}{||\bm{d}_t||_2}.\label{eq:normalized_attention_distance_vector}
\end{equation}
The number of dimensions in $d_k$ is the number of the vocabulary (fixed length) for Word iAT, while the dimension of words in a sentence (variable length) for Attention iAT.
The dimensionality of $d_t$ in Attention iAT is much smaller compared to Word iAT.\footnote{This normalization is done on a sentence-by-sentence basis, so it does not matter that the dimension is varies ($T << V$).}
We define perturbation $r(\bm{\alpha}_t)$ for attention to the $t$-th word with trainable parameters $\bm{\alpha}_t = (\alpha_{t,k})_{t=1}^{T} \in \mathbb{R}^{T}$ and the normalized difference vector of the attention $\tilde{\bm{d}}_t$ as follows:
\begin{equation}
    r(\bm{\alpha}_t) = \bm{\alpha}^{\top}_t \cdot \tilde{\bm{d}}_t.
\end{equation}
By combining $\bm{\alpha}_t$ for all $t$, we can calculate perturbation $\bm{r}(\bm{\alpha})$ for the sentence:
\begin{equation}
    \bm{r}(\bm{\alpha}) = \{r(\bm{\alpha}_t)\}_{t=1}^{T}.
\end{equation}
Then, similar to $X_{\tilde{\bm{a}} + \bm{r}}$ in Eq.~\ref{eq:argmax}, we introduce $X_{\bm{\tilde{a}} + \bm{r}(\bm{\alpha})}$ and seek the worst-case weights of the difference vectors that maximize the loss functions as follows:
\begin{equation}
    \bm{r}_{\tt iAT} = \argmax_{\bm{\alpha}:||\bm{\alpha}|| \le \epsilon} \mathcal{L}(X_{\tilde{\bm{a}} + \bm{r}({\bm{\alpha}})}, \bm{y}; \hat{\bm{\theta}}).
\end{equation}

Contrary to Attention iAT, in Word iAT, the difference $d_{t,k}$ in Eq.~\ref{eq:dist_attention_in_sentence} is defined as the distance between the $t$-th word in a sentence and the $k$-th word in the vocabulary in the word embedding space.
Based on the distance, Word iAT determines the direction of perturbation for the $t$-th word as a linear sum of the word directional vectors in the vocabulary.
In contrast, Attention iAT does not compute the distance to word embeddings in the vocabulary.
Instead, this technique computes the difference in attention to other words in the sentence and determines the direction of the perturbation.
The adversarial perturbation of Attention iAT, defined in this way, works to increase the difference in attention to each word.
We call this process in Attention iAT as \textit{attention difference enhancement}.
Owing to the process, Attention iAT improves the interpretability of attention and contributes to the performance of the model's prediction.
The detail discussions are shared in Section~\ref{sec:discussion_attention_at_attention_iat_word_iat}.

For computational efficiency, we calculate the interpretable adversarial perturbation by applying the same approximation method as in Eq.~\ref{eq:fast_gradient_method}:
\begin{equation}
    \bm{r}_{\tt iAT} = \epsilon \frac{\bm{g}}{||\bm{g}||_2}, \mathrm{where}~~ \bm{g} = \nabla_{\bm{\alpha}} \mathcal{L}(X_{\tilde{\bm{a}} + \bm{r}({\bm{\alpha}})}, \bm{y}; \hat{\bm{\theta}}).
\end{equation}
Then, similar to Eq.~\ref{eq:alpha_plus_r_at}, we construct a perturbated example for attention score $\tilde{\bm{a}}$:
\begin{equation}
    \tilde{\bm{a}}_{\tt iadv} = \tilde{\bm{a}} + \bm{r}_{\tt iAT}.
\end{equation}

\subsection{Training a Model with Adversarial Training}
At each training step, we generate adversarial perturbation in the current model.
To this end, we define the loss function for adversarial training as follows:
\begin{equation}
    \tilde{\mathcal{L}} = \underbrace{\mathcal{L}(X_{\tilde{\bm{a}}}, \bm{y}; \bm{\theta})}_{\substack{\text{The loss from} \\ \text{unmodified examples}}} + \underbrace{\lambda \mathcal{L}(X_{ \tilde{\bm{a}}_{\tt ADV}}, \bm{y}; \bm{\theta})}_{\substack{\text{The loss from} \\ \text{its adversarial examples}}},\label{eq:overall_loss}
\end{equation}
where $\lambda$ is the coefficient that controls the balance between two loss functions.
Note that $X_{\tilde{\bm{a}}_{\tt ADV}}$ can be $X_{\tilde{\bm{a}}_{\tt adv}}$ for Attention AT or $X_{\tilde{\bm{a}}_{\tt iadv}}$ for Attention iAT.

\section{Experiments}
In this section, we describe the evaluation tasks and datasets, the details of the models, and the evaluation criteria.

\subsection{Tasks and Datasets}\label{sec:dataset}
We evaluated the proposed techniques using the open benchmark tasks (i.e., four BCs, four QAs, and two NLIs) used in Jain and Wallace~\cite{jain2019attention}.
In our experiment, we added MultiNLI~\cite{williams2017broad} as an additional NLI task for more detailed analysis (see the details in Appendix~\ref{sec:tasks_and_datasets}).
Table~\ref{tab:dataset} presents the statistics for all datasets.
We split the dataset into a training set, a validation set, and a test set.\footnote{Jain and Wallace~\cite{jain2019attention} split the dataset into only a training set and a test set, so we did not get the same results.}
We performed preprocessing, including tokenization with \href{https://spacy.io/}{spaCy}\footnote{spaCy · Industrial-strength Natural Language Processing in Python \url{https://spacy.io/}}, mapping out vocabulary words to a special $\texttt{<unk>}$ token, and mapping all words with numeric characters to $\texttt{qqq}$ in the same manner as Jain and Wallace~\cite{jain2019attention}.

\subsection{Model Settings}\label{sec:models}
We compared the two proposed training techniques to four conventional training techniques.
They were implemented using the same model architecture as described in Section~\ref{sec:common_model_architecture}.
Following Jain and Wallace~\cite{jain2019attention}, we used bi-directional long short-term memory (LSTM)~\cite{hochreiter1997long} as the BiRNN-based encoder, including \textbf{Enc}, $\mathbf{Enc}_P$, and $\mathbf{Enc}_Q$.
A total of six training techniques were evaluated in the experiments:
\begin{itemize}
    \item \textbf{Vanilla}~\cite{jain2019attention}: a model with attention mechanisms trained without the use of AT.
    \item \textbf{Word AT}~\cite{miyato2016adversarial}: word embeddings trained with AT.
    \item \textbf{Word iAT}~\cite{sato2018interpretable}: word embeddings trained with iAT.
    \item \textbf{Attention RP}: attention to the word embeddings is trained with random perturbation (RP).
    \item \textbf{Attention AT} (\textbf{Proposed}): attention to the word embeddings is trained with AT.
    \item \textbf{Attention iAT} (\textbf{Proposed}): attention to the word embeddings is trained with iAT.
\end{itemize}
We implemented the training techniques above using the AllenNLP library with Interpret~\cite{gardner2018allennlp,Wallace2019AllenNLP}.
Through the experiments, we set the hyper-parameter $\lambda = 1$ related to AT or iAT in Eq.~\ref{eq:overall_loss}.
To ensure a fair comparison of the training techniques, we followed the configurations (e.g., initialization of word embedding, hidden size of the encoder, optimizer settings) used in the literature~\cite{jain2019attention} (see the details in Appendix~\ref{sec:implementation_detail}).

Note that while Jain and Wallace~\cite{jain2019attention} used a test set to adjust the model's hyper-parameters, we used a validation set.
In adversarial training, the Allentune library~\cite{dodge2019show} was used to adjust hyper-parameter $\epsilon$, and we report the test scores for the model with the highest validation score.

\subsection{Evaluation Criteria}
First, we compared the prediction performance of each model for each task.
As an evaluation metric of the prediction performance, we used the F1 score\footnote{The F1 score is a metric that harmonizes precision and recall.. Therefore, this score takes both false positives and false negatives into account.}, accuracy, and the micro-F1 score for the BC, QA, and NLI, respectively, as in \cite{jain2019attention}.

Next, we compared how the attention weights obtained through the proposed AT-based technique agreed with the importance of words calculated by the gradients~\cite{simonyan2013deep}.
To evaluate the agreement, we compared the Pearson's correlations between the attention weights and the word importance of the gradient-based method.
In \cite{jain2019attention}, the Kendall tau, which represents rank correlation, was used to evaluate the relationship between attention and the word importance obtained by the gradients.
Recently, however, it has been pointed out that rank correlations often misrepresent the relationship between the two due to the noise in the order of the low rankings~\cite{mohankumar2020towards}; we concurred with this, so we used Pearson's correlations.

Finally, we compared the effects of perturbation size $\epsilon$ of AT on the validation performance of the BC, QA, and NLI tasks with a fixed $\lambda = 1$.
We randomly choose the value of $\epsilon$ in the 0--30 range and ran the training 100 times.
The configurations in~\cite{miyato2016adversarial, sato2018interpretable} were $\epsilon = 5$ for Word AT and $\epsilon = 15$ for Word iAT.

\section{Results}
\begin{table*}[t]
\centering
\caption{
    Comparison of prediction performance and the Pearson's correlation coefficients (Corr.) between the attention weight and the word importance of the gradient-based method. 
    We used the same test metrics as in \cite{jain2019attention}: binary classification (BC), question answering (QA), and natural language inference (NLI). 
    As an evaluation metrics of the prediction performance, we used the F1 score (F1), accuracy (Acc.), and the micro-F1 (Micro-F1) score for BC, QA, and NLI, respectively.
}
\begin{minipage}{\linewidth}
    %{\footnotesize
\centering
\subcaption{
    Binary classification (BC)
}
\begin{tabular}{@{}lrrrrrrrr@{}}
\toprule
\multicolumn{1}{c}{\multirow{2}{*}{Model}} & \multicolumn{2}{c}{SST}                                 & \multicolumn{2}{c}{IMDB}                                & \multicolumn{2}{c}{20News}                              & \multicolumn{2}{c}{AGNews}                              \\ \cmidrule(r){2-3} \cmidrule(lr){4-5} \cmidrule(lr){6-7} \cmidrule(l){8-9}
\multicolumn{1}{c}{}                       & \multicolumn{1}{c}{F1 [\%]} & \multicolumn{1}{c}{Corr.} & \multicolumn{1}{c}{F1 [\%]} & \multicolumn{1}{c}{Corr.} & \multicolumn{1}{c}{F1 [\%]} & \multicolumn{1}{c}{Corr.} & \multicolumn{1}{c}{F1 [\%]} & \multicolumn{1}{c}{Corr.} \\ \cmidrule(r){1-1} \cmidrule(lr){2-2} \cmidrule(lr){3-3} \cmidrule(lr){4-4} \cmidrule(lr){5-5} \cmidrule(lr){6-6} \cmidrule(lr){7-7} \cmidrule(lr){8-8} \cmidrule(l){9-9}
Vanilla~\cite{jain2019attention}           & 79.27                       & 0.652                     & 88.77                       & 0.788                     & 95.05                       & 0.891                     & 95.27                       & 0.822                     \\ \cmidrule(r){1-1} \cmidrule(lr){2-3} \cmidrule(lr){4-5} \cmidrule(lr){6-7} \cmidrule(l){8-9}
Word AT~\cite{miyato2016adversarial}       & 79.61                       & 0.647                     & 89.65                       & 0.838                     & 95.11                       & 0.892                     & 95.59                       & 0.813                     \\
Word iAT~\cite{sato2018interpretable}      & 79.57                       & 0.643                     & 89.64                       & 0.839                     & 95.14                       & 0.893                     & 95.62                       & 0.809                     \\ \cmidrule(r){1-1} \cmidrule(lr){2-3} \cmidrule(lr){4-5} \cmidrule(lr){6-7} \cmidrule(l){8-9}
Attention RP               & 81.90                       & 0.531                     & 89.79                       & 0.628                     & 96.09                       & 0.883                     & 96.08                       & 0.792                     \\ \cmidrule(r){1-1} \cmidrule(lr){2-3} \cmidrule(lr){4-5} \cmidrule(lr){6-7} \cmidrule(l){8-9}
Attention AT (\textbf{Proposed})                        & 81.72                       & 0.852                     & 90.00                       & 0.819                     & \textbf{96.69}              & 0.868                     & 96.12                       & 0.835                     \\
Attention iAT (\textbf{Proposed})                       & \textbf{82.20}              & \textbf{0.876}            & \textbf{90.21}              & \textbf{0.861}            & 96.58                       & \textbf{0.897}            & \textbf{96.19}              & \textbf{0.891}            \\ \bottomrule
\end{tabular}
%\vspace{-1mm}

\label{tab:bc_score}
%}

% \cmidrule(r){1-1} \cmidrule(lr){2-2} \cmidrule(lr){3-3} \cmidrule(lr){4-4} \cmidrule(lr){5-5} \cmidrule(lr){6-6} \cmidrule(lr){7-7} \cmidrule(lr){8-8} \cmidrule(l){9-9}
    \vspace{5mm}
\end{minipage} \\
\begin{minipage}{\linewidth}
    %{\footnotesize
\centering
\subcaption{
    Question answering (QA)
}
\begin{tabular}{@{}lrrrrrrrr@{}}
\toprule
\multicolumn{1}{c}{\multirow{3}{*}{Model}} & \multicolumn{2}{c}{CNN news}                                                                     & \multicolumn{6}{c}{bAbI}                                                                                                                                                          \\ \cmidrule(r){2-3} \cmidrule(l){4-9}
\multicolumn{1}{c}{}                       & \multicolumn{1}{c}{\multirow{2}{*}{Acc. [\%]}} & \multicolumn{1}{c}{\multirow{2}{*}{Corr.}} & \multicolumn{2}{c}{Task 1}                                & \multicolumn{2}{c}{Task 2}                                & \multicolumn{2}{c}{Task 3}                                \\ \cmidrule(lr){4-5} \cmidrule(lr){6-7} \cmidrule(l){8-9}
\multicolumn{1}{c}{}                       & \multicolumn{1}{c}{}                           & \multicolumn{1}{c}{}                       & \multicolumn{1}{c}{Acc. [\%]} & \multicolumn{1}{c}{Corr.} & \multicolumn{1}{c}{Acc. [\%]} & \multicolumn{1}{c}{Corr.} & \multicolumn{1}{c}{Acc. [\%]} & \multicolumn{1}{c}{Corr.} \\ \cmidrule(r){1-1} \cmidrule(lr){2-2} \cmidrule(lr){3-3} \cmidrule(lr){4-4} \cmidrule(lr){5-5} \cmidrule(lr){6-6} \cmidrule(lr){7-7} \cmidrule(lr){8-8} \cmidrule(l){9-9}
Vanilla~\cite{jain2019attention}           & 64.95                                          & 0.765                                      & 99.90                         & 0.714                     & 45.10                         & 0.459                     & 52.00                         & 0.387                     \\ \cmidrule(r){1-1} \cmidrule(lr){2-3} \cmidrule(lr){4-5} \cmidrule(lr){6-7} \cmidrule(l){8-9}
Word AT~\cite{miyato2016adversarial}       & 65.67                                          & 0.779                                      & \textbf{100.00}               & 0.797                     & 79.50                         & 0.657                     & 55.10                         & 0.439                     \\
Word iAT~\cite{sato2018interpretable}      & 65.66                                          & 0.776                                      & 99.90                         & 0.798                     & 79.80                         & 0.658                     & 54.90                         & 0.437                     \\ \cmidrule(r){1-1} \cmidrule(lr){2-3} \cmidrule(lr){4-5} \cmidrule(lr){6-7} \cmidrule(l){8-9}
Attention RP                        & 65.78                                          & 0.614                                      & \textbf{100.00}               & 0.592                     & 80.60                         & 0.584                     & 55.35                         & 0.373                     \\ \cmidrule(r){1-1} \cmidrule(lr){2-3} \cmidrule(lr){4-5} \cmidrule(lr){6-7} \cmidrule(l){8-9}
Attention AT (\textbf{Proposed})                        & 65.93                                          & 0.771                                      & \textbf{100.00}               & 0.807                     & 82.30                         & 0.632                     & 56.00                         & 0.514                     \\
Attention iAT (\textbf{Proposed})                       & \textbf{66.17}                                 & \textbf{0.784}                             & \textbf{100.00}               & \textbf{0.821}            & \textbf{85.40}                & \textbf{0.710}            & \textbf{57.10}                & \textbf{0.589}            \\ \bottomrule
\end{tabular}

\label{tab:qa_score}
%}

    \vspace{5mm}
\end{minipage} \\
\begin{minipage}{\linewidth}
    %{\footnotesize
\centering
\subcaption{
    Natural language inference (NLI)
}
\begin{tabular}{@{}lrrrrrr@{}}
\toprule
\multicolumn{1}{c}{\multirow{3}{*}{Model}} & \multicolumn{2}{c}{SNLI}                                                                        & \multicolumn{4}{c}{Multi NLI}                                                                                                        \\ \cmidrule(r){2-3} \cmidrule(l){4-7}
\multicolumn{1}{c}{}                       & \multicolumn{1}{l}{\multirow{2}{*}{Micro-F1 [\%]}} & \multicolumn{1}{l}{\multirow{2}{*}{Corr.}} & \multicolumn{3}{c}{Micro-F1 [\%]}                                                       & \multicolumn{1}{l}{\multirow{2}{*}{Corr.}} \\ \cmidrule(lr){4-6}
\multicolumn{1}{c}{}                       & \multicolumn{1}{l}{}                               & \multicolumn{1}{l}{}                       & \multicolumn{1}{c}{Avg.} & \multicolumn{1}{c}{Matched} & \multicolumn{1}{c}{Mismatched} & \multicolumn{1}{l}{}                       \\ \cmidrule(r){1-1} \cmidrule(lr){2-2} \cmidrule(lr){3-3} \cmidrule(lr){4-4} \cmidrule(lr){5-5} \cmidrule(lr){6-6} \cmidrule(l){7-7}
Vanilla~\cite{jain2019attention}           & 78.64                                              & 0.764                                      & 60.26                    & 59.80                       & 60.71                          & 0.541                                      \\ \cmidrule(r){1-1} \cmidrule(lr){2-3} \cmidrule(lr){4-5} \cmidrule(l){6-7}
Word AT~\cite{miyato2016adversarial}       & 79.03                                              & 0.812                                      & 60.72                    & 60.58                       & 60.86                          & 0.601                                      \\
Word iAT~\cite{sato2018interpretable}      & 79.12                                              & 0.815                                      & 60.73                    & 60.59                       & 60.87                          & 0.603                                      \\ \cmidrule(r){1-1} \cmidrule(lr){2-3} \cmidrule(lr){4-5} \cmidrule(l){6-7}
Attention RP                               & 79.23                                              & 0.569                                      & 60.97                    & 61.02                       & 60.91                          & 0.547                                      \\ \cmidrule(r){1-1} \cmidrule(lr){2-3} \cmidrule(lr){4-5} \cmidrule(l){6-7}
Attention AT (\textbf{Proposed})                        & 79.19                                              & 0.792                                      & 61.17                    & 61.20                       & \textbf{61.13}                 & 0.626                                      \\
Attention iAT (\textbf{Proposed})                       & \textbf{79.32}                                     & \textbf{0.818}                             & \textbf{61.34}           & \textbf{61.75}              & 60.93                          & \textbf{0.668}                             \\ \bottomrule
\end{tabular}

\label{tab:nli_score}
%}
\end{minipage}
\label{tab:classification_scores}
\end{table*}

In this section, we share the results of the experiments.
Table~\ref{tab:classification_scores} presents the prediction performance and the Pearson's correlations between the attention weight for the words and word importance calculated from the model gradient.
The most significant results are shown in bold.

\subsection{Comparison of Prediction Performance}

In terms of prediction performance, the model that applied the proposed Attention AT/iAT demonstrated a clear advantage over the model without AT (as shown in Vanilla~\cite{jain2019attention}) as well as other AT-based techniques (Word AT~\cite{miyato2016adversarial} and Word iAT~\cite{sato2018interpretable}). 
The proposed technique achieved the best results in almost all benchmarks.
For 20News and AGNews in the BC and bAbI task 1 in QA, the conventional techniques, including the Vanilla model, were sufficiently accurate (the score was higher than 95\%), so the performance improvement of the proposed techniques to the tasks was limited to some extent.
Meanwhile, Attention AT/iAT contributed to solid performance improvements in other complicated tasks.

\begin{figure*}[t]
    \begin{minipage}{\linewidth}
        \centering
        \includegraphics[width=0.9\linewidth]{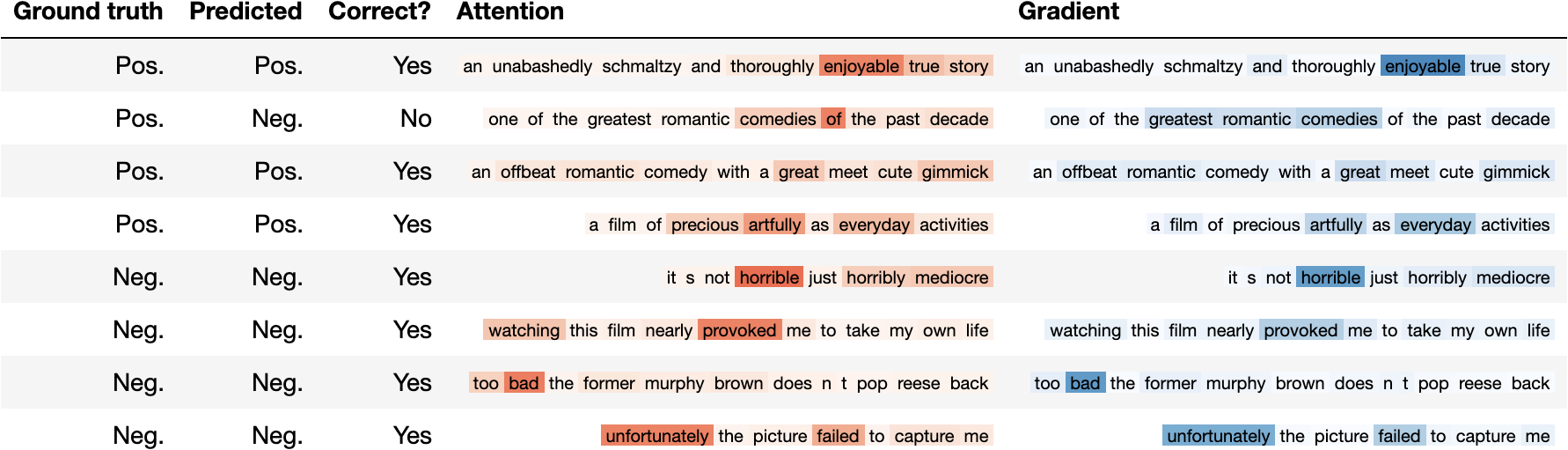}
        \subcaption{Vanilla}
    \end{minipage}

    \begin{minipage}{\linewidth}
    \vspace{2mm}
    \end{minipage}

    \begin{minipage}{\linewidth}
        \centering
        \includegraphics[width=0.9\linewidth]{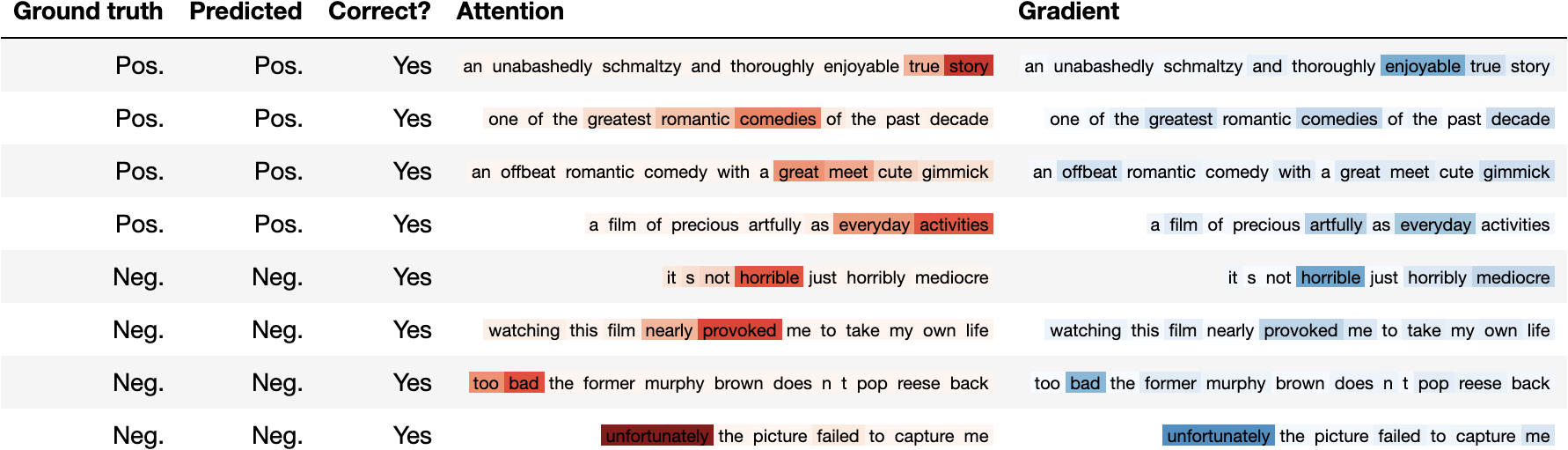}
        \subcaption{(\textbf{Proposed}) Attention AT}
    \end{minipage}
    
    \begin{minipage}{\linewidth}
    \vspace{2mm}
    \end{minipage}
    
    \begin{minipage}{\linewidth}
        \centering
        \includegraphics[width=0.9\linewidth]{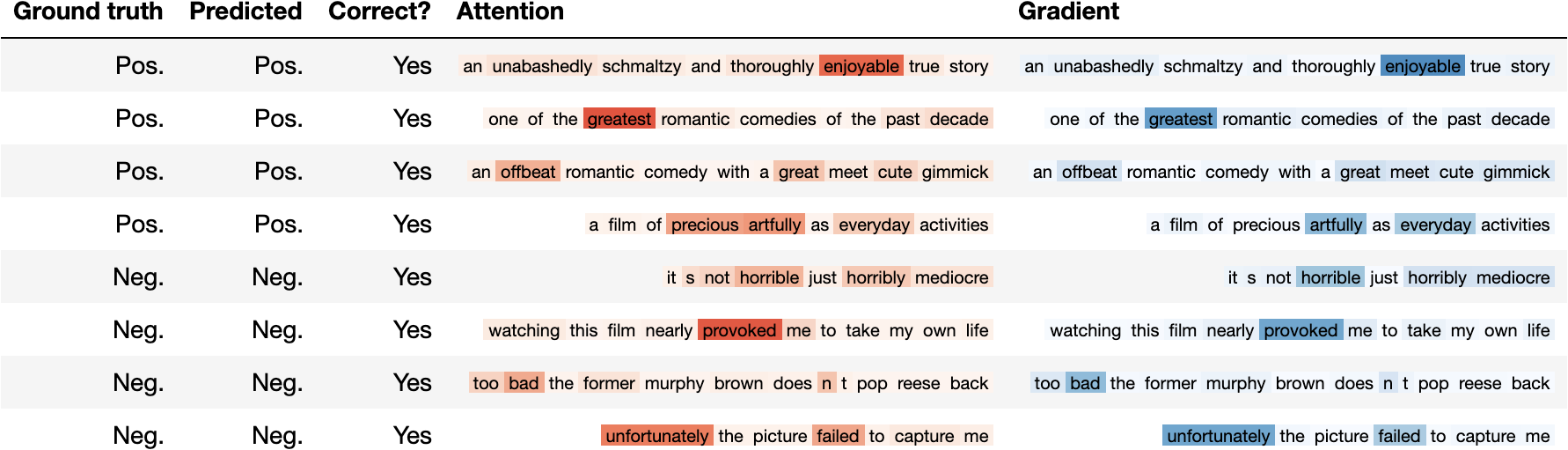}
        \subcaption{(\textbf{Proposed}) Attention iAT}
    \end{minipage}
    \caption{Visualization of attention weight for each word and word importance calculated by gradients on the SST dataset. 
    Best viewed in color.
    The models that apply the proposed Attention AT/iAT give clearer attention.
    In terms of word importance, Attention iAT has more similar attention-based and gradient-based results than the other methods.
    }
    \label{fig:attention_gradient_sst}
\end{figure*}

\subsection{Comparison of Correlation between Attention Weights and Gradients on Word Importance}
In terms of model interpretability, the attention to the words obtained with the Attention AT/iAT techniques notably correlated with the importance of the word as determined by the gradients.
Attention iAT demonstrated the highest correlation among the techniques in all benchmarks.
Figure~\ref{fig:attention_gradient_sst} visualizes the attention weight for each word and gradient-based word importance in the SST test dataset.
Attention AT yielded clearer attention compared to the Vanilla model or Attention iAT.
Specifically, Attention AT tended to strongly focus attention on a few words.
Regarding the correlation of word importance based on attention weights and gradient-based word importance, Attention iAT demonstrated higher similarities than the other models.

\begin{figure*}[t]
    \begin{minipage}{0.325\linewidth}
        \centering
        \includegraphics[width=\linewidth]{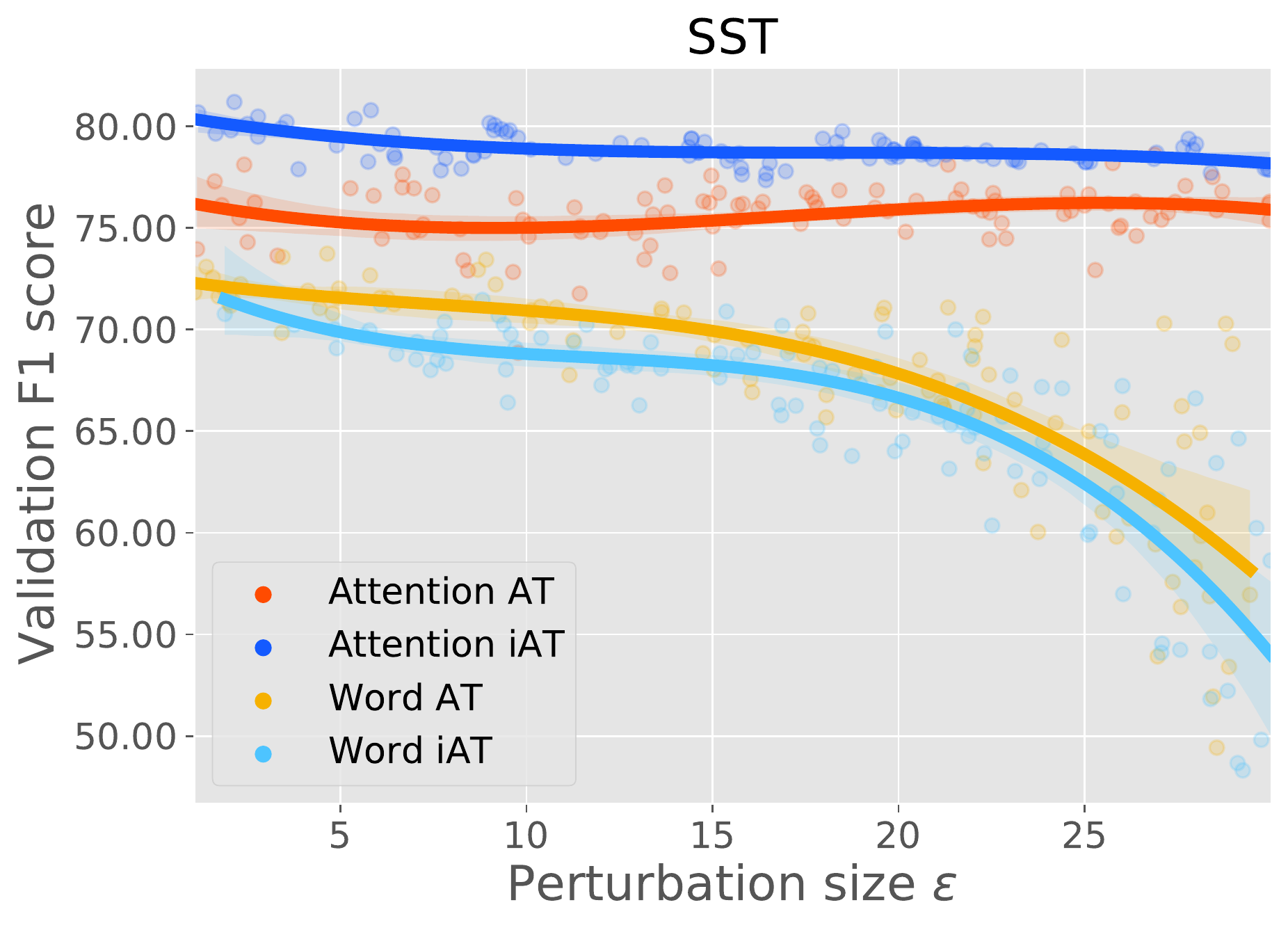}
        \subcaption{SST (BC)}
    \end{minipage}
    \begin{minipage}{0.325\linewidth}
        \centering
        \includegraphics[width=\linewidth]{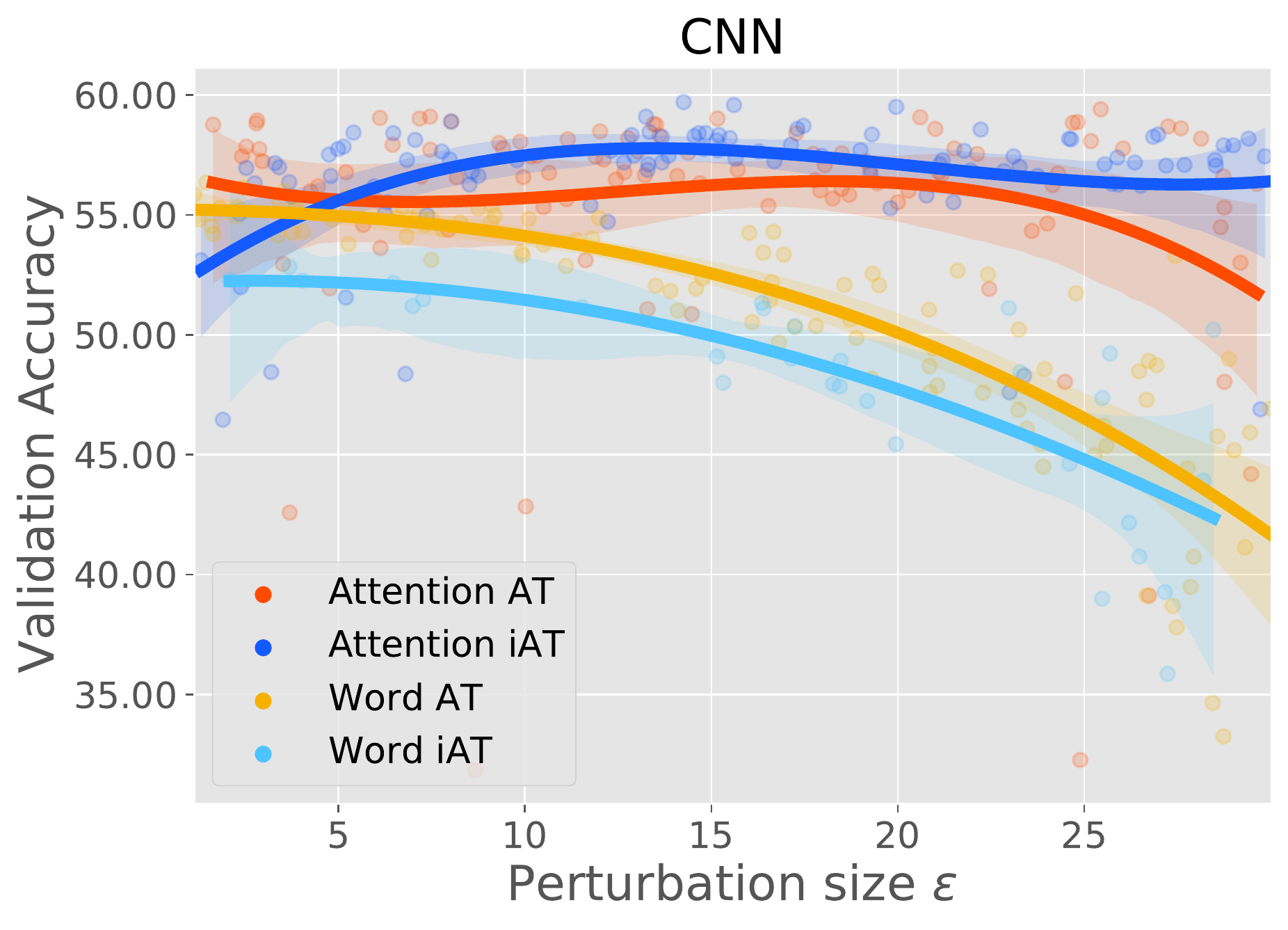}
        \subcaption{CNN news (QA)}
    \end{minipage}
    \begin{minipage}{0.325\linewidth}
        \centering
        \includegraphics[width=\linewidth]{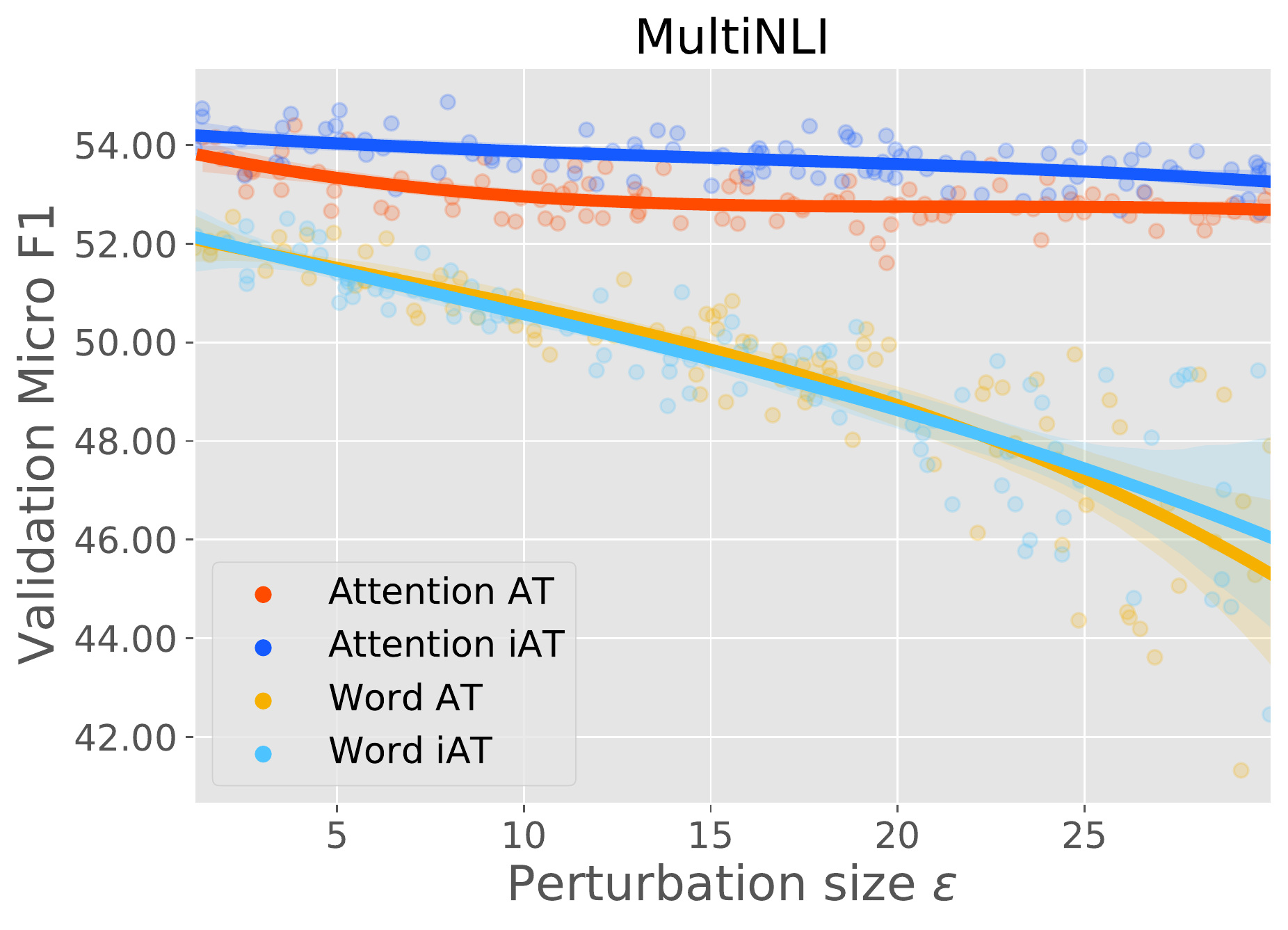}
        \subcaption{Multi NLI (NLI)}
    \end{minipage}
    \caption{
        The effect of perturbation size $\epsilon$ on the validation performance.
        We observed that the model with Attention AT/iAT maintained an almost constant prediction performance even when the perturbation size increased.
        Compared to Word AT/iAT, The model with the proposed  techniques, Attention AT/iAT, was more robust with a large $\epsilon$.
    }
    \label{fig:score_xi}
\end{figure*}

\subsection{Effects of Perturbation Size}

Figure~\ref{fig:score_xi} shows the effect of the perturbation size $\epsilon$ on the validation performance of SST (BC), CNN news (QA), and Multi NLI (NLI) with a fixed $\lambda = 1$.
We observed that the performances of the conventional Word AT/iAT techniques deteriorated according to the increase in the perturbation size; meanwhile, our Attention AT/iAT techniques maintained almost the same prediction performance.
We observed similar trends in other datasets as described in Section~\ref{sec:dataset}. 

\section{Discussion}
\subsection{Comparison of Adversarial Training for Attention Mechanisms and Word Embedding}

Attention AT/iAT is based on our hypothesis that attention is more important in finding significant words in document processing than the word embeddings themselves. 
Therefore, we sought to achieve prediction performance and model interpretability by introducing AT to the attention mechanisms. 
We confirmed that the application of AT to the attention mechanisms (Attention AT/iAT) was more effective than word embedding (Word AT/iAT) and supports the correctness of our hypothesis, as shown in Table~\ref{tab:classification_scores}.
In particular, the Attention iAT technique was not only more accurate in its model than the Word AT/iAT techniques but also demonstrated a higher correlation with the importance of the words predicted based on the gradient.

As shown in Figure~\ref{fig:attention_gradient_sst}, Attention AT tended to display more attention to the sentence than the Vanilla model.
The results showed that training with adversarial perturbations to the attention mechanism allowed for cleaner attention without changing word meanings or grammatical functions.
Furthermore, we confirmed that the proposed Attention AT/iAT techniques were more robust regarding the variation of perturbation size $\epsilon$ than conventional Word AT/iAT, as shown in Figure~\ref{fig:score_xi}.
Although it is difficult to directly compare perturbations to attention and word embedding because of the difference in the range of the perturbation size to the part, the model that added perturbations to attention behaved robustly even when the perturbations were relatively large.

\subsection{Comparison of Random Perturbations and Adversarial Perturbations}

Attention RP demonstrated better prediction performance than Word AT/iAT.
The results revealed that augmentation for the attention mechanism is very effective, even with simple random noise.
In contrast, the correlations between the attention weight for the word and the gradient-based word importance were significantly reduced, as shown in Table~\ref{tab:classification_scores}.
We consider that Attention RP is successful in learning robust discriminative boundaries through random perturbation and improving to the desired classification performance.
However, as the gradient is smoothed out by the perturbation around the (supervised) data points, the correlation with the word importance by the gradient is considered to be degraded.
In other words, Attention RP can achieve a certain level of classification performance, but it does not lead to which words are useful from their gradients.

\subsection{Comparison of Attention AT, Attention iAT, and Word iAT}\label{sec:discussion_attention_at_attention_iat_word_iat}

In the experiments, Attention iAT showed a better performance compared Attention AT in the prediction performance and the correlation with the gradient-based word importance.
Attention iAT exploits the difference in the attention weight of each word in a sentence to determine adversarial perturbations.
Because the norm of the difference in attention weight (as shown in Eq.~\ref{eq:normalized_attention_distance_vector}) is normalized to one, adversarial perturbations in attention mechanisms will make these differences clear, especially in the case of sentences with a small difference in the attention to each word.
That is, even in situations where there is little difference in attention between each element of $\bm{d}_t = (d_{t,1}, d_{t,2}, \cdots, d_{t,T})$, the difference is amplified by Eq.~\ref{eq:normalized_attention_distance_vector}.
Therefore, even for the same perturbation size $||\bm{\alpha}|| < \epsilon$, more effective perturbations $\bm{r}(\bm{\alpha})$ weighted by $\tilde{\bm{d}}_t$ were successfully obtained for each word.
However, in the case of sentences where there was originally a difference in the clear attention to each word, the regularization of $\bm{d}_t$ in Eq.~\ref{eq:normalized_attention_distance_vector} had practically no effect because it did not change their ratio nearly as much.
Thus, we posit that the Attention iAT technique enhances the effectiveness of AT applied to attention mechanisms by generating effective perturbations for each word.

The Attention iAT technique was inspired by Word iAT.
Word iAT generates perturbations in the direction that maximizes the loss function while restricting the direction of the perturbation to become a linear combination of the direction of word embedding in the vocabulary.
Word iAT indirectly improves the interpretability of the model by indicating which words in the vocabulary to which the perturbation is similar.
However, we are confident that Attention iAT is a direct improvement in the interpretability of the model, because it can show more clearly which words to pay attention.
This is owing to the attention difference enhancement process described in Eq.~\ref{eq:normalized_attention_distance_vector}.
Thus, the proposed techniques are highly effective in that they lead to a more substantive improvement in interpretability.

\subsection{Limitations}

Our proposal is a general-purpose robust training technique for DNN models, which are commonly used for NLP tasks.
Therefore, we have chosen here an RNN with an attention mechanism that has been put to practical use~\cite{jain2019attention}. 
For this reason, models such as BERT~\cite{devlin2019bert} that deal with self-attention were outside the scope of this study, and will be the subject of future work. 
We also did not deal with tasks (such as machine translation) that were not used in the literature~\cite{jain2019attention} as baselines.
Additionally, for the same reason in as \cite{serrano2019attention}, we did not consider the variants of attention mechanisms, such as bi-attentive architecture~\cite{parikh2016decomposable}, multi-headed architecture~\cite{vaswani2017attention}, because they could have different interpretability properties.

As an extension of AT, virtual adversarial training (VAT), which is a semi-supervised training technique, was proposed in \cite{miyato2016adversarial, miyato2018virtual}.
Based on VAT, the proposed technique can be expected to improve accuracy by using unlabeled datasets.

\section{Conclusion}
We proposed robust and interpretable attention training techniques that exploit AT.
In the experiments with various NLP tasks, we confirmed that AT for attention mechanisms achieves better performance than techniques using AT for word embedding in terms of the prediction performance and the interpretability of the model.
Specifically, the Attention iAT technique introduced adversarial perturbations that emphasized differences in the importance of words in a sentence and combined high accuracy with interpretable attention, which was more strongly correlated with the gradient-based method of word importance.
The proposed technique could be applied to various models and NLP tasks.
This paper provides strong support and motivation for utilizing AT with attention mechanisms in NLP tasks.

In the experiment, we demonstrated the effectiveness of the proposed techniques for RNN models that are reported to be vulnerable to attention mechanisms, but we will confirm the effectiveness of the proposed technique for large language models with attention mechanisms such as Transformer~\cite{vaswani2017attention} or BERT~\cite{devlin2019bert} in the future.
Because the proposed techniques are model-independent and general techniques for attention mechanisms, we can expect they will improve predictability and the interpretability for language models.

\appendices
\section{Tasks and Dataset}\label{sec:tasks_and_datasets}

\subsection{Binary Classification}
The following datasets were used for evaluation.
The Stanford Sentiment Treebank (SST)~\cite{socher2013recursive}\footnote{\url{https://nlp.stanford.edu/sentiment/trainDevTestTrees_PTB.zip}} was used to ascertain positive or negative sentiment from a sentence.
IMDB Large Movie Reviews (IMDB)~\cite{maas2011learning}\footnote{\url{https://s3.amazonaws.com/text-datasets/imdb_full.pkl}}$^{,}$\footnote{\url{https://s3.amazonaws.com/text-datasets/imdb_word_index.json}} was used to identify positive or negative sentiment from movie reviews.
20 Newsgroups (20News)~\cite{lang1995newsweeder}\footnote{\url{https://ndownloader.figshare.com/files/5975967}} were used to ascertain the topic of news articles as either baseball (set as a negative label) or hockey (set as a positive label).
The AG News (AGNews)~\cite{zhang2015character}\footnote{The dataset can be found on Xiang Zhang's \href{https://drive.google.com/drive/u/0/folders/0Bz8a_Dbh9Qhbfll6bVpmNUtUcFdjYmF2SEpmZUZUcVNiMUw1TWN6RDV3a0JHT3kxLVhVR2M}{Google Drive}.} was used to identify the topic of news articles as either world  (set as a negative label) or business (set as a positive label).

\subsection{Question Answering}
The following datasets were used for evaluation.
The CNN news article corpus (CNN news)~\cite{hermann2015teaching}\footnote{The dataset can be found on Deep Mind Q\&A \href{https://drive.google.com/uc?export=download&id=0BwmD_VLjROrfTTljRDVZMFJnVWM}{Google Drive}.} was used to identify answer entities from a paragraph.
The bAbI dataset (bAbI)~\cite{weston2015towards}\footnote{\url{https://research.fb.com/downloads/babi/}} contains 20 different question-answer tasks, and we considered three tasks: (task 1) basic factoid question answered with a single supporting fact, (task 2) factoid question answered with two supporting facts, and (task 3) factoid question answered with three supporting facts.
The model was trained for each task.

\subsection{Natural Language Inference}
The following datasets were used for evaluation.
The Stanford Natural Language Inference (SNLI)~\cite{bowman2015large}\footnote{\url{https://nlp.stanford.edu/projects/snli/snli_1.0.zip}} is used to identify whether a hypothesis sentence entails, contradicts, or is neutral concerning a given premise sentence.
Multi-Genre NLI (MultiNLI)~\cite{williams2017broad}\footnote{\url{https://www.nyu.edu/projects/bowman/multinli/multinli_1.0.zip}} uses the same format as SNLI and is comparable in size, but it includes a more diverse range of text, as well as an auxiliary test set for cross-genre transfer evaluation.

\section{Implementation Detail}\label{sec:implementation_detail}
For all datasets, we either used pretrained GloVe~\cite{pennington2014glove} or fastText~\cite{bojanowski2017enriching} word embedding with 300 dimensions except the bAbI dataset.
For the bAbI dataset, we trained 50 dimensional word embeddings from scratch during training.
We used a one-layer LSTM as the encoder with a hidden size of 64 for the bAbI dataset and 256 for the other datasets.
All models were regularized using $L_2$ regularization ($10^{-5}$) applied to all parameters.
We trained the model using the maximum likelihood loss utilizing the Adam~\cite{kingma2014adam} optimizer with a learning rate of 0.001.

\section*{Acknowledgment}
We would like to appreciate the editors and anonymous reviewers for their helpful feedback.
We also thank Quan Huu Cap and Mahmoud Daif for feedback and fruitful discussions.

\bibliographystyle{IEEEtran}
\bibliography{references}

\begin{thebibliography}{10}
\providecommand{\url}[1]{#1}
\csname url@samestyle\endcsname
\providecommand{\newblock}{\relax}
\providecommand{\bibinfo}[2]{#2}
\providecommand{\BIBentrySTDinterwordspacing}{\spaceskip=0pt\relax}
\providecommand{\BIBentryALTinterwordstretchfactor}{4}
\providecommand{\BIBentryALTinterwordspacing}{\spaceskip=\fontdimen2\font plus
\BIBentryALTinterwordstretchfactor\fontdimen3\font minus
  \fontdimen4\font\relax}
\providecommand{\BIBforeignlanguage}[2]{{%
\expandafter\ifx\csname l@#1\endcsname\relax
\typeout{** WARNING: IEEEtran.bst: No hyphenation pattern has been}%
\typeout{** loaded for the language `#1'. Using the pattern for}%
\typeout{** the default language instead.}%
\else
\language=\csname l@#1\endcsname
\fi
#2}}
\providecommand{\BIBdecl}{\relax}
\BIBdecl

\bibitem{bahdanau2014neural}
\BIBentryALTinterwordspacing
D.~Bahdanau, K.~Cho, and Y.~Bengio, ``Neural machine translation by jointly
  learning to align and translate,'' \emph{CoRR preprint arXiv:1409.0473},
  2014. [Online]. Available: \url{https://arxiv.org/abs/1409.0473}
\BIBentrySTDinterwordspacing

\bibitem{lin2017structured}
\BIBentryALTinterwordspacing
Z.~Lin, M.~Feng, C.~N. dos Santos, M.~Yu, B.~Xiang, B.~Zhou, and Y.~Bengio, ``A
  structured self-attentive sentence embedding,'' in \emph{Proc. of the 5th
  International Conference on Learning Representations, {ICLR}, Conference
  Track Proceedings}, 2017. [Online]. Available:
  \url{https://openreview.net/forum?id=BJC_jUqxe&noteId=BJC_jUqxe}
\BIBentrySTDinterwordspacing

\bibitem{wang2016attention}
\BIBentryALTinterwordspacing
Y.~Wang, M.~Huang, and L.~Zhao, ``Attention-based {LSTM} for aspect-level
  sentiment classification,'' in \emph{Proc. of the 2016 Conference on
  Empirical Methods in Natural Language Processing}, ser. Association for
  Computational Linguistics (ACL), 2016, pp. 606--615. [Online]. Available:
  \url{http://dx.doi.org/10.18653/v1/D16-1058}
\BIBentrySTDinterwordspacing

\bibitem{he2016character}
\BIBentryALTinterwordspacing
X.~He and D.~Golub, ``Character-level question answering with attention,'' in
  \emph{Proc. of the 2016 Conference on Empirical Methods in Natural Language
  Processing}, ser. Association for Computational Linguistics (ACL), 2016, pp.
  1598--1607. [Online]. Available: \url{http://dx.doi.org/10.18653/v1/D16-1166}
\BIBentrySTDinterwordspacing

\bibitem{parikh2016decomposable}
\BIBentryALTinterwordspacing
A.~Parikh, O.~T{\"a}ckstr{\"o}m, D.~Das, and J.~Uszkoreit, ``A decomposable
  attention model for natural language inference,'' in \emph{Proc. of the 2016
  Conference on Empirical Methods in Natural Language Processing}, ser.
  Association for Computational Linguistics (ACL), 2016, pp. 2249--2255.
  [Online]. Available: \url{http://dx.doi.org/10.18653/v1/D16-1244}
\BIBentrySTDinterwordspacing

\bibitem{luong2015effective}
\BIBentryALTinterwordspacing
T.~Luong, H.~Pham, and C.~D. Manning, ``Effective approaches to attention-based
  neural machine translation,'' in \emph{Proc. of the 2015 Conference on
  Empirical Methods in Natural Language Processing}, ser. Association for
  Computational Linguistics (ACL), 2015, pp. 1412--1421. [Online]. Available:
  \url{http://dx.doi.org/10.18653/v1/D15-1166}
\BIBentrySTDinterwordspacing

\bibitem{rush2015neural}
\BIBentryALTinterwordspacing
A.~M. Rush, S.~Chopra, and J.~Weston, ``A neural attention model for
  abstractive sentence summarization,'' in \emph{Proc. of the 2015 Conference
  on Empirical Methods in Natural Language Processing}, ser. Association for
  Computational Linguistics (ACL), 2015, pp. 379--389. [Online]. Available:
  \url{http://dx.doi.org/10.18653/v1/D15-1044}
\BIBentrySTDinterwordspacing

\bibitem{vaswani2017attention}
\BIBentryALTinterwordspacing
A.~Vaswani, N.~Shazeer, N.~Parmar, J.~Uszkoreit, L.~Jones, A.~N. Gomez,
  {\L}.~Kaiser, and I.~Polosukhin, ``Attention is all you need,'' in
  \emph{Proc. of the 30th International Conference on Neural Information
  Processing Systems}, 2017, pp. 5998--6008. [Online]. Available:
  \url{https://arxiv.org/abs/1706.03762}
\BIBentrySTDinterwordspacing

\bibitem{szegedy2013intriguing}
\BIBentryALTinterwordspacing
C.~Szegedy, W.~Zaremba, I.~Sutskever, J.~Bruna, D.~Erhan, I.~Goodfellow, and
  R.~Fergus, ``Intriguing properties of neural networks,'' in \emph{2nd
  International Conference on Learning Representations, {ICLR}, Conference
  Track Proceedings}, 2013. [Online]. Available:
  \url{https://arxiv.org/abs/1312.6199}
\BIBentrySTDinterwordspacing

\bibitem{jain2019attention}
\BIBentryALTinterwordspacing
S.~Jain and B.~C. Wallace, ``Attention is not explanation,'' in \emph{Proc. of
  the 2019 Conference of the North American Chapter of the Association for
  Computational Linguistics: Human Language Technologies, Volume 1 (Long and
  Short Papers)}, ser. Association for Computational Linguistics (ACL), 2019,
  pp. 3543--3556. [Online]. Available:
  \url{http://dx.doi.org/10.18653/v1/N19-1357}
\BIBentrySTDinterwordspacing

\bibitem{liu2016delving}
\BIBentryALTinterwordspacing
Y.~Liu, X.~Chen, C.~Liu, and D.~Song, ``Delving into transferable adversarial
  examples and black-box attacks,'' \emph{CoRR preprint arXiv:1611.02770},
  2016. [Online]. Available: \url{https://arxiv.org/abs/1611.02770}
\BIBentrySTDinterwordspacing

\bibitem{liu2019roberta}
\BIBentryALTinterwordspacing
Y.~Liu, M.~Ott, N.~Goyal, J.~Du, M.~Joshi, D.~Chen, O.~Levy, M.~Lewis,
  L.~Zettlemoyer, and V.~Stoyanov, ``Roberta: A robustly optimized bert
  pretraining approach,'' \emph{CoRR preprint arXiv:1907.11692}, 2019.
  [Online]. Available: \url{https://arxiv.org/abs/1907.11692}
\BIBentrySTDinterwordspacing

\bibitem{lan2020albert}
\BIBentryALTinterwordspacing
Z.~Lan, M.~Chen, S.~Goodman, K.~Gimpel, P.~Sharma, and R.~Soricut, ``Albert: A
  lite bert for self-supervised learning of language representations,'' in
  \emph{International Conference on Learning Representations}, 2020. [Online].
  Available: \url{https://openreview.net/forum?id=H1eA7AEtvS}
\BIBentrySTDinterwordspacing

\bibitem{tay2020efficient}
\BIBentryALTinterwordspacing
Y.~Tay, M.~Dehghani, D.~Bahri, and D.~Metzler, ``Efficient transformers: A
  survey,'' \emph{CoRR preprint arXiv:2009.06732}, 2020. [Online]. Available:
  \url{https://arxiv.org/abs/2009.06732}
\BIBentrySTDinterwordspacing

\bibitem{goodfellow2014explaining}
\BIBentryALTinterwordspacing
I.~J. Goodfellow, J.~Shlens, and C.~Szegedy, ``Explaining and harnessing
  adversarial examples,'' in \emph{3rd International Conference on Learning
  Representations, {ICLR}, Conference Track Proceedings}, 2014. [Online].
  Available: \url{http://arxiv.org/abs/1412.6572}
\BIBentrySTDinterwordspacing

\bibitem{shaham2018understanding}
\BIBentryALTinterwordspacing
U.~Shaham, Y.~Yamada, and S.~Negahban, ``Understanding adversarial training:
  Increasing local stability of supervised models through robust
  optimization,'' \emph{Neurocomputing}, vol. 307, pp. 195--204, 2018.
  [Online]. Available: \url{https://doi.org/10.1016/j.neucom.2018.04.027}
\BIBentrySTDinterwordspacing

\bibitem{socher2013recursive}
\BIBentryALTinterwordspacing
R.~Socher, A.~Perelygin, J.~Wu, J.~Chuang, C.~D. Manning, A.~Ng, and C.~Potts,
  ``Recursive deep models for semantic compositionality over a sentiment
  treebank,'' in \emph{Proc. of the 2013 Conference on Empirical Methods in
  Natural Language Processing}, ser. Association for Computational Linguistics
  (ACL), 2013, pp. 1631--1642. [Online]. Available:
  \url{https://www.aclweb.org/anthology/D13-1170/}
\BIBentrySTDinterwordspacing

\bibitem{miyato2016adversarial}
\BIBentryALTinterwordspacing
T.~Miyato, A.~M. Dai, and I.~Goodfellow, ``Adversarial training methods for
  semi-supervised text classification,'' in \emph{5th International Conference
  on Learning Representations, {ICLR}, Conference Track Proceedings}, 2016.
  [Online]. Available: \url{https://openreview.net/forum?id=r1X3g2_xl}
\BIBentrySTDinterwordspacing

\bibitem{sato2018interpretable}
\BIBentryALTinterwordspacing
M.~Sato, J.~Suzuki, H.~Shindo, and Y.~Matsumoto, ``Interpretable adversarial
  perturbation in input embedding space for text,'' in \emph{Proc. of the 27th
  International Joint Conference on Artificial Intelligence}, ser. AAAI Press,
  2018, pp. 4323--4330. [Online]. Available:
  \url{https://dl.acm.org/doi/10.5555/3304222.3304371}
\BIBentrySTDinterwordspacing

\bibitem{yasunaga2018robust}
\BIBentryALTinterwordspacing
M.~Yasunaga, J.~Kasai, and D.~Radev, ``Robust multilingual part-of-speech
  tagging via adversarial training,'' in \emph{Proc. of the 2018 Conference of
  the North American Chapter of the Association for Computational Linguistics:
  Human Language Technologies, Volume 1 (Long Papers)}, ser. Association for
  Computational Linguistics (ACL), 2018, pp. 976--986. [Online]. Available:
  \url{http://dx.doi.org/10.18653/v1/N18-1089}
\BIBentrySTDinterwordspacing

\bibitem{wang2018robust}
\BIBentryALTinterwordspacing
Y.~Wang and M.~Bansal, ``Robust machine comprehension models via adversarial
  training,'' in \emph{Proc. of the 2018 Conference of the North American
  Chapter of the Association for Computational Linguistics: Human Language
  Technologies, Volume 2 (Short Papers)}, ser. Association for Computational
  Linguistics (ACL), 2018, pp. 575--581. [Online]. Available:
  \url{http://dx.doi.org/10.18653/v1/N18-2091}
\BIBentrySTDinterwordspacing

\bibitem{li2016understanding}
\BIBentryALTinterwordspacing
J.~Li, W.~Monroe, and D.~Jurafsky, ``Understanding neural networks through
  representation erasure,'' \emph{CoRR preprint arXiv:1612.08220}, 2016.
  [Online]. Available: \url{https://arxiv.org/abs/1612.08220}
\BIBentrySTDinterwordspacing

\bibitem{tsipras2019robustness}
\BIBentryALTinterwordspacing
D.~Tsipras, S.~Santurkar, L.~Engstrom, A.~Turner, and A.~Madry, ``Robustness
  may be at odds with accuracy,'' in \emph{International Conference on Learning
  Representations}, no. 2019, 2019. [Online]. Available:
  \url{https://openreview.net/forum?id=SyxAb30cY7}
\BIBentrySTDinterwordspacing

\bibitem{itazuri2019adversarially}
\BIBentryALTinterwordspacing
T.~Itazuri, Y.~Fukuhara, H.~Kataoka, and S.~Morishima, ``What do adversarially
  robust models look at?'' \emph{CoRR preprint arXiv:1905.07666}, 2019.
  [Online]. Available: \url{https://arxiv.org/abs/1905.07666}
\BIBentrySTDinterwordspacing

\bibitem{zhang2019interpreting}
\BIBentryALTinterwordspacing
T.~Zhang and Z.~Zhu, ``Interpreting adversarially trained convolutional neural
  networks,'' in \emph{International Conference on Machine Learning}.\hskip 1em
  plus 0.5em minus 0.4em\relax PMLR, 2019, pp. 7502--7511. [Online]. Available:
  \url{http://proceedings.mlr.press/v97/zhang19s.html}
\BIBentrySTDinterwordspacing

\bibitem{simonyan2013deep}
\BIBentryALTinterwordspacing
K.~Simonyan, A.~Vedaldi, and A.~Zisserman, ``Deep inside convolutional
  networks: Visualising image classification models and saliency maps,'' in
  \emph{2nd International Conference on Learning Representations, {ICLR},
  Workshop Track Proceedings}, 2013. [Online]. Available:
  \url{https://arxiv.org/abs/1312.6034}
\BIBentrySTDinterwordspacing

\bibitem{wang2016theoretical}
\BIBentryALTinterwordspacing
B.~Wang, J.~Gao, and Y.~Qi, ``A theoretical framework for robustness of (deep)
  classifiers against adversarial examples,'' \emph{CoRR preprint
  arXiv:1612.00334}, 2016. [Online]. Available:
  \url{https://arxiv.org/abs/1612.00334}
\BIBentrySTDinterwordspacing

\bibitem{liu2020robust}
\BIBentryALTinterwordspacing
K.~Liu, X.~Liu, A.~Yang, J.~Liu, J.~Su, S.~Li, and Q.~She, ``A robust
  adversarial training approach to machine reading comprehension,'' in
  \emph{Proc. of the AAAI Conference on Artificial Intelligence}, vol.~34,
  no.~05, 2020, pp. 8392--8400. [Online]. Available:
  \url{https://doi.org/10.1609/aaai.v34i05.6357}
\BIBentrySTDinterwordspacing

\bibitem{barham2019interpretable}
\BIBentryALTinterwordspacing
S.~Barham and S.~Feizi, ``Interpretable adversarial training for text,''
  \emph{CoRR preprint arXiv:1905.12864}, 2019. [Online]. Available:
  \url{https://arxiv.org/abs/1905.12864}
\BIBentrySTDinterwordspacing

\bibitem{zhu2019retrieval}
\BIBentryALTinterwordspacing
Q.~Zhu, L.~Cui, W.-N. Zhang, F.~Wei, and T.~Liu, ``Retrieval-enhanced
  adversarial training for neural response generation,'' in \emph{Proc. of the
  57th Annual Meeting of the Association for Computational Linguistics}.\hskip
  1em plus 0.5em minus 0.4em\relax Association for Computational Linguistics,
  2019, pp. 3763--3773. [Online]. Available:
  \url{http://dx.doi.org/10.18653/v1/P19-1366}
\BIBentrySTDinterwordspacing

\bibitem{srivastava2014dropout}
\BIBentryALTinterwordspacing
N.~Srivastava, G.~Hinton, A.~Krizhevsky, I.~Sutskever, and R.~Salakhutdinov,
  ``Dropout: a simple way to prevent neural networks from overfitting,''
  \emph{The journal of machine learning research}, vol.~15, no.~1, pp.
  1929--1958, 2014. [Online]. Available:
  \url{https://jmlr.org/papers/v15/srivastava14a.html}
\BIBentrySTDinterwordspacing

\bibitem{ioffe2015batch}
\BIBentryALTinterwordspacing
S.~Ioffe and C.~Szegedy, ``Batch normalization: Accelerating deep network
  training by reducing internal covariate shift,'' in \emph{International
  Conference on Machine Learning}, 2015, pp. 448--456. [Online]. Available:
  \url{http://proceedings.mlr.press/v37/ioffe15.html}
\BIBentrySTDinterwordspacing

\bibitem{iyyer2015deep}
\BIBentryALTinterwordspacing
M.~Iyyer, V.~Manjunatha, J.~Boyd-Graber, and H.~Daum{\'e}~III, ``Deep unordered
  composition rivals syntactic methods for text classification,'' in
  \emph{Proc. of the 53rd Annual Meeting of the Association for Computational
  Linguistics and the 7th International Joint Conference on Natural Language
  Processing (Volume 1: Long Papers)}, 2015, pp. 1681--1691. [Online].
  Available: \url{http://dx.doi.org/10.3115/v1/P15-1162}
\BIBentrySTDinterwordspacing

\bibitem{shimada2016document}
\BIBentryALTinterwordspacing
D.~Shimada, R.~Kotani, and H.~Iyatomi, ``Document classification through
  image-based character embedding and wildcard training,'' in \emph{Proc. of
  IEEE International Conference on Big Data}.\hskip 1em plus 0.5em minus
  0.4em\relax IEEE, 2016, pp. 3922--3927. [Online]. Available:
  \url{https://doi.org/10.1109/BigData.2016.7841067}
\BIBentrySTDinterwordspacing

\bibitem{maas2011learning}
\BIBentryALTinterwordspacing
A.~L. Maas, R.~E. Daly, P.~T. Pham, D.~Huang, A.~Y. Ng, and C.~Potts,
  ``Learning word vectors for sentiment analysis,'' in \emph{Proc. of the 49th
  Annual Meeting of the Association for Computational Linguistics: Human
  Language Technologies}, ser. Association for Computational Linguistics (ACL),
  vol.~1, 2011, pp. 142--150. [Online]. Available:
  \url{https://www.aclweb.org/anthology/P11-1015/}
\BIBentrySTDinterwordspacing

\bibitem{lang1995newsweeder}
\BIBentryALTinterwordspacing
K.~Lang, ``Newsweeder: Learning to filter netnews,'' in \emph{Machine Learning
  Proceedings}.\hskip 1em plus 0.5em minus 0.4em\relax Elsevier, 1995, pp.
  331--339. [Online]. Available:
  \url{https://doi.org/10.1016/B978-1-55860-377-6.50048-7}
\BIBentrySTDinterwordspacing

\bibitem{zhang2015character}
\BIBentryALTinterwordspacing
X.~Zhang, J.~Zhao, and Y.~LeCun, ``Character-level convolutional networks for
  text classification,'' in \emph{Proc. of the 28th International Conference on
  Neural Information Processing Systems}, ser. MIT Press, vol.~1, 2015, pp.
  649--657. [Online]. Available:
  \url{https://dl.acm.org/doi/10.5555/2969239.2969312}
\BIBentrySTDinterwordspacing

\bibitem{hermann2015teaching}
\BIBentryALTinterwordspacing
K.~M. Hermann, T.~Kocisky, E.~Grefenstette, L.~Espeholt, W.~Kay, M.~Suleyman,
  and P.~Blunsom, ``Teaching machines to read and comprehend,'' in \emph{Proc.
  of the 28th International Conference on Neural Information Processing
  Systems}, ser. MIT Press, vol.~1, 2015, pp. 1693--1701. [Online]. Available:
  \url{https://dl.acm.org/doi/10.5555/2969239.2969428}
\BIBentrySTDinterwordspacing

\bibitem{weston2015towards}
\BIBentryALTinterwordspacing
J.~Weston, A.~Bordes, S.~Chopra, and T.~Mikolov, ``Towards {AI}-complete
  question answering: {A} set of prerequisite toy tasks,'' in \emph{4th
  International Conference on Learning Representations, {ICLR}, Conference
  Track Proceedings}, 2016. [Online]. Available:
  \url{https://arxiv.org/abs/1502.05698}
\BIBentrySTDinterwordspacing

\bibitem{bowman2015large}
\BIBentryALTinterwordspacing
S.~Bowman, G.~Angeli, C.~Potts, and C.~D. Manning, ``A large annotated corpus
  for learning natural language inference,'' in \emph{Proc. of the 2015
  Conference on Empirical Methods in Natural Language Processing}, ser.
  Association for Computational Linguistics (ACL), 2015, pp. 632--642.
  [Online]. Available: \url{http://dx.doi.org/10.18653/v1/D15-1075}
\BIBentrySTDinterwordspacing

\bibitem{williams2017broad}
\BIBentryALTinterwordspacing
A.~Williams, N.~Nangia, and S.~R. Bowman, ``A broad-coverage challenge corpus
  for sentence understanding through inference,'' in \emph{Proc. of the 2018
  Conference of the North American Chapter of the Association for Computational
  Linguistics: Human Language Technologies (Long Papers)}, ser. Association for
  Computational Linguistics (ACL), 2017. [Online]. Available:
  \url{http://dx.doi.org/10.18653/v1/N18-1101}
\BIBentrySTDinterwordspacing

\bibitem{hochreiter1997long}
\BIBentryALTinterwordspacing
S.~Hochreiter and J.~Schmidhuber, ``Long short-term memory,'' \emph{Neural
  Computation}, vol.~9, no.~8, pp. 1735--1780, 1997. [Online]. Available:
  \url{https://doi.org/10.1162/neco.1997.9.8.1735}
\BIBentrySTDinterwordspacing

\bibitem{gardner2018allennlp}
\BIBentryALTinterwordspacing
M.~Gardner, J.~Grus, M.~Neumann, O.~Tafjord, P.~Dasigi, N.~Liu, M.~Peters,
  M.~Schmitz, and L.~Zettlemoyer, ``Allennlp: A deep semantic natural language
  processing platform,'' \emph{CoRR preprint arXiv:1803.07640}, 2018. [Online].
  Available: \url{http://dx.doi.org/10.18653/v1/W18-2501}
\BIBentrySTDinterwordspacing

\bibitem{Wallace2019AllenNLP}
\BIBentryALTinterwordspacing
E.~Wallace, J.~Tuyls, J.~Wang, S.~Subramanian, M.~Gardner, and S.~Singh,
  ``{AllenNLP Interpret}: A framework for explaining predictions of {NLP}
  models,'' in \emph{Proc. of the 2019 Conference on Empirical Methods in
  Natural Language Processing and the 9th International Joint Conference on
  Natural Language Processing (EMNLP-IJCNLP): System Demonstrations}, ser.
  Association for Computational Linguistics (ACL), 2019, pp. 7--12. [Online].
  Available: \url{http://dx.doi.org/10.18653/v1/D19-3002}
\BIBentrySTDinterwordspacing

\bibitem{dodge2019show}
\BIBentryALTinterwordspacing
J.~Dodge, S.~Gururangan, D.~Card, R.~Schwartz, and N.~A. Smith, ``Show your
  work: Improved reporting of experimental results,'' in \emph{Proc. of the
  2019 Conference on Empirical Methods in Natural Language Processing and the
  9th International Joint Conference on Natural Language Processing
  (EMNLP-IJCNLP)}, ser. Association for Computational Linguistics (ACL), 2019,
  pp. 2185--2194. [Online]. Available:
  \url{http://dx.doi.org/10.18653/v1/D19-1224}
\BIBentrySTDinterwordspacing

\bibitem{mohankumar2020towards}
\BIBentryALTinterwordspacing
A.~K. Mohankumar, P.~Nema, S.~Narasimhan, M.~M. Khapra, B.~V. Srinivasan, and
  B.~Ravindran, ``Towards transparent and explainable attention models,'' in
  \emph{Proc. of the 58th Annual Meeting of the Association for Computational
  Linguistics}, ser. Association for Computational Linguistics (ACL), 2020, pp.
  4206--4216. [Online]. Available:
  \url{http://dx.doi.org/10.18653/v1/2020.acl-main.387}
\BIBentrySTDinterwordspacing

\bibitem{devlin2019bert}
\BIBentryALTinterwordspacing
J.~Devlin, M.-W. Chang, K.~Lee, and K.~Toutanova, ``Bert: Pre-training of deep
  bidirectional transformers for language understanding,'' in \emph{Proc. of
  the 2019 Conference of the North American Chapter of the Association for
  Computational Linguistics: Human Language Technologies, Volume 1 (Long and
  Short Papers)}, ser. Association for Computational Linguistics (ACL), 2019,
  pp. 4171--4186. [Online]. Available:
  \url{http://dx.doi.org/10.18653/v1/N19-1423}
\BIBentrySTDinterwordspacing

\bibitem{serrano2019attention}
\BIBentryALTinterwordspacing
S.~Serrano and N.~A. Smith, ``Is attention interpretable?'' in \emph{Proc. of
  the 57th Annual Meeting of the Association for Computational Linguistics},
  ser. Association for Computational Linguistics (ACL), 2019, pp. 2931--2951.
  [Online]. Available: \url{http://dx.doi.org/10.18653/v1/P19-1282}
\BIBentrySTDinterwordspacing

\bibitem{miyato2018virtual}
\BIBentryALTinterwordspacing
T.~Miyato, S.-i. Maeda, M.~Koyama, and S.~Ishii, ``Virtual adversarial
  training: a regularization method for supervised and semi-supervised
  learning,'' \emph{IEEE transactions on pattern analysis and machine
  intelligence}, vol.~41, no.~8, pp. 1979--1993, 2018. [Online]. Available:
  \url{https://doi.org/10.1109/TPAMI.2018.2858821}
\BIBentrySTDinterwordspacing

\bibitem{pennington2014glove}
\BIBentryALTinterwordspacing
J.~Pennington, R.~Socher, and C.~Manning, ``Glove: Global vectors for word
  representation,'' in \emph{Proc. of the 2014 Conference on Empirical Methods
  in Natural Language Processing}, ser. Association for Computational
  Linguistics (ACL), 2014, pp. 1532--1543. [Online]. Available:
  \url{http://dx.doi.org/10.3115/v1/D14-1162}
\BIBentrySTDinterwordspacing

\bibitem{bojanowski2017enriching}
\BIBentryALTinterwordspacing
P.~Bojanowski, E.~Grave, A.~Joulin, and T.~Mikolov, ``Enriching word vectors
  with subword information,'' \emph{Transactions of the Association for
  Computational Linguistics}, vol.~5, pp. 135--146, 2017. [Online]. Available:
  \url{http://dx.doi.org/10.1162/tacl_a_00051}
\BIBentrySTDinterwordspacing

\bibitem{kingma2014adam}
\BIBentryALTinterwordspacing
D.~P. Kingma and J.~Ba, ``Adam: A method for stochastic optimization,''
  \emph{CoRR preprint arXiv:1412.6980}, 2014. [Online]. Available:
  \url{https://arxiv.org/abs/1412.6980}
\BIBentrySTDinterwordspacing

\end{thebibliography}

\EOD

\end{document}